\documentclass[a4paper,fleqn]{cas-dc}

\usepackage[numbers]{natbib}
\usepackage{amsmath}
\usepackage{graphicx}
\usepackage{booktabs} 
\usepackage{algorithm}
\usepackage{algpseudocode}
\usepackage{multirow}
\usepackage{subcaption}
\usepackage{tabularx} 
\usepackage{algorithm}
\usepackage{algpseudocode}

\begin{document}
\let\WriteBookmarks\relax
\def\floatpagepagefraction{1}
\def\textpagefraction{.001}

\shorttitle{LGFNet: Local-Global Fusion Network}
\shortauthors{ et~al.}

\title [mode = title]{LGFNet: Local-Global Fusion Network with Fidelity Gap Delta Learning for Multi-Source Aerodynamics}

\author[1]{Qinye Zhu}

\author[1]{Yu Xiang}
\cormark[1] 
\ead{jcxiang@uestc.edu.cn}
\cortext[1]{Corresponding author}

\author[1]{Jun Zhang}
\author[1]{Wenyong Wang}

\affiliation[1]{organization={School of Computer Science and Engineering, University of Electronic Science and Technology of China},
                city={Chengdu},
                country={China}}

\begin{abstract}
The precise fusion of computational fluid dynamic (CFD) data, wind tunnel tests data, and flight tests data in aerodynamic area is essential for obtaining comprehensive knowledge of both localized flow structures and global aerodynamic trends across the entire flight envelope. However, existing methodologies often struggle to balance high-resolution local fidelity with wide-range global dependency, leading to either a loss of sharp discontinuities or an inability to capture long-range topological correlations.
We propose Local-Global Fusion Network (LGFNet) for multi-scale feature decomposition to extract this dual-natured aerodynamic knowledge. To this end, LGFNet combines a spatial perception layer that integrates a sliding window mechanism with a relational reasoning layer based on self-attention, simultaneously reinforcing the continuity of fine-grained local features (e.g., shock waves) and capturing long-range flow information. Furthermore, the fidelity gap delta learning (FGDL) strategy is proposed to treat CFD data as a "low-frequency carrier" to explicitly approximate nonlinear discrepancies. This approach prevents unphysical smoothing while inheriting the foundational physical trends from the simulation baseline. Experiments demonstrate that LGFNet achieves state-of-the-art (SOTA) performance in both accuracy and uncertainty reduction across diverse aerodynamic scenarios.
\end{abstract}



\maketitle

\section{Introduction}
In contemporary aerospace engineering, the acquisition of precise aerodynamic data is the cornerstone of aircraft design, flight envelope expansion, and performance evaluation \cite{Kumar2025, Wang2025}. Moreover, reliable aerodynamic modeling is indispensable for accurate parameter identification and the development of robust flight control systems \cite{Chauhan2017, Ding2025}. At present, the main sources for obtaining aerodynamic data of aircraft are Computational Fluid Dynamic (CFD) simulation, wind tunnel tests, and flight tests. While CFD can generate a large volume of data, their accuracy is often compromised due to model assumptions and limitations in computational resources. Wind tunnel tests, on the other hand, can provide relatively accurate aerodynamic data in a controlled environment, but they are costly, time-consuming, and typically require physical models, which may restrict their application in certain scenarios. Flight tests offer aerodynamic data with high fidelity, often serving as the ultimate validation standard. However, flight tests are also subject to significant limitations, including high costs, long durations, and difficulty in meeting boundary conditions \cite{Tang2023,Poloczek2017,Hu2025}.

Given the distinct limitations of each method described above, relying on a single source is inadequate for high-precision aerodynamic analysis. However, since these sources characterize the same aircraft, they possess inherent physical correlations. Consequently, it is essential to employ Multi-Source Data Fusion (MDF) to combine the large-scale and wide-range CFD data with the high-precision test data. This approach effectively mitigates the predictive uncertainties of low-fidelity models, simultaneously balancing accuracy improvement with cost reduction across the entire flight envelope \cite{Qiu2024}. Through MDF, aerodynamic accuracy, cost control and system optimization can be integrated into a unified design framework, which is conducive to significantly improve the overall aircraft development level \cite{He2014}.

To bridge the fidelity gap, traditional surrogate models like Kriging and Co-Kriging \cite{JeromeSacks1989,Forrester2007} are widely used for robust, unbiased aerodynamic data estimation. To improve computational efficiency, Hierarchical Kriging (HK) \cite{Han2012} calibrates a low-fidelity global trend using an independent discrepancy model. However, despite their rigorous mathematical foundations, these methods often face scalability bottlenecks and struggle to capture high-frequency non-linearities, such as sharp shock waves, in dense datasets.

Machine learning (ML) and deep learning (DL) have emerged as pivotal tools for handling complex data, owing to their formidable capabilities in non-linear feature extraction \cite{Brunton_2020}. In particular, the application of ML and DL to MDF has yielded significant advancements in enhancing data accuracy and mitigating acquisition costs.

Machine learning (ML) is a key approach for addressinging issues of weight allocation and sample imbalance in data fusion \cite{Wang2023, Cui2024}. It is notable that purely data-driven tree models have demonstrated potential in nonlinear regression. For instance, Lin et al. proposed a Grafted XGBoost model, which innovatively constructs a baseline trend using low-fidelity data and subsequently "grafts" corrections derived from high-fidelity residuals, effectively enhancing prediction robustness under sparse sample conditions \cite{Lin2024}.

Deep learning (DL) has fundamentally advanced the modeling of high-dimensional flow fields. Various architectures, including CNNs \cite{Liao2021,HU2023108198}, DNNs \cite{He2020,Zuo2023,2022_Li}, BNNs \cite{Vaiuso2024,Chen2025}, Composite Neural Networks \cite{WANG2025111707,Ning2024}, GANs \cite{WU2022470,WANG202262,Hu2025}, and MTL \cite{HU2022108369}, have been extensively deployed for aerodynamic predictions. Among these, CNNs are particularly prominent due to their superior capability in extracting spatial hierarchical features from grid-based flow fields via local connections and parameter sharing \cite{Bhatnagar2019,Nils2020}.

Although the aforementioned methods—including convolutional ones—excel in local feature extraction, they often suffer from "texture bias" \cite{Azad2021}, making it difficult to effectively capture global topological structures and long-range dependency information essential for complex flow fields.

To overcome the limitations of traditional convolutional networks in capturing long-range dependencies, the Self-Attention Mechanism \cite{Vaswani2017} has been introduced into fluid dynamics. By establishing global context through connection weights, this mechanism offers a novel paradigm for aerodynamic prediction. Recent studies have successfully incorporated various attention architectures—including feature-level attention \cite{Dong2025}, physics-informed cross-attention \cite{WangY2025}, distance attention \cite{Fang2025}, and geometry-aware operator learning (e.g., ArGEnT) \cite{Chen2026ArGEnT}—to enhance aerodynamic modeling capabilities.

However, despite the excellence of self-attention mechanisms in global feature capture, challenges remain in aerodynamic data fusion tasks. Standard softmax-based self-attention fundamentally acts as a low-pass filter \cite{Wi2025}. While this property is highly beneficial for smoothing out high-frequency measurement noise and interpolation artifacts, it inadvertently causes pure attention architectures to struggle with extracting fine-grained, high-frequency local features, such as sharp shock wave discontinuities \cite{ZHOU2025128242}. Moreover, the quadratic complexity ($O(N^2)$) associated with full attention calculation imposes high demands on computational resources, limiting large-scale engineering applications.

To address the aforementioned challenges in feature representation, we propose Local-Global Fusion Network (LGFNet). The core innovation of this model lies in its integrated local-global fusion architecture, which is further enhanced by the Fidelity Gap Delta Learning (FGDL) strategy to achieve superior multi-source aerodynamic data synergy. The specific contributions are summarized as follows:

\begin{enumerate}
    \item A specialized framework specifically for aerodynamic data fusion is proposed, consisting of three interdependent functional layers: the spatial perception layer (SPL), the relational reasoning layer (RRL), and the feature synthesis layer (FSL). This structure is capable of simultaneously extracting and integrating both local and global aerodynamic information.
    \item A methodology that integrates a sliding-window method, a self-attention mechanism, and a residual-driven synthesis process is introduced. The sliding window reinforces the coverage and continuity of local variations, the self-attention mechanism acts as a global low-pass filter to dynamically weigh data relationships and suppress interpolation noise, and the FGDL strategy reformulates the fusion task into a residual learning process to preserve the underlying aerodynamic trends.
    \item Experimental Validation and SOTA Performance: The efficacy of LGFNet is demonstrated through extensive experiments on surface pressure distribution fusion (Scenario 1) and high-dimensional aerodynamic coefficient fusion (Scenario 2). Results across the RAE2822 airfoil and CARDC aircraft datasets show that the model achieves state-of-the-art (SOTA) performance in both predictive accuracy and uncertainty reduction.
\end{enumerate}

\section{Methodology}
\subsection{Relevant definitions and Overview}
Our work aims to leverage the rich physical trend information inherent in low-fidelity data to complete or calibrate sparse high-fidelity data. 

For the convenience of description in the following text, we describe the mathematical symbols used in this article in Table \ref{tab:definitions}.

\begin{table}[htbp]
    \centering
    \caption{Relevant Definitions and Notations}
    \label{tab:definitions}
    \renewcommand{\arraystretch}{1.3} 
    \begin{tabularx}{\columnwidth}{l X} 
        \toprule
        \textbf{Symbol} & \textbf{Description} \\
        \midrule
        $\mathbf{x}$ & Input aerodynamic state vector  \\
        $\mathbf{y}$ & Output aerodynamic response vector  \\
        $\mathbf{d} = [\mathbf{x}, \mathbf{y}]$ & Unified aerodynamic data vector  \\
        $N_L, N_H$ & The quantity of samples in low-fidelity (CFD) and high-fidelity datasets  \\
        $\mathcal{D}_L, \mathcal{D}_H$ & Original low-fidelity and high-fidelity datasets  \\
        $\mathcal{T}_L, \mathcal{T}_H$ & Aligned datasets generated via sliding window transformation  \\
        $L, S_{tr}$ & Length and stride size of the sliding window  \\
        \bottomrule
    \end{tabularx}
\end{table}

\begin{figure*}
	\centering
	\includegraphics[width=0.95\textwidth]{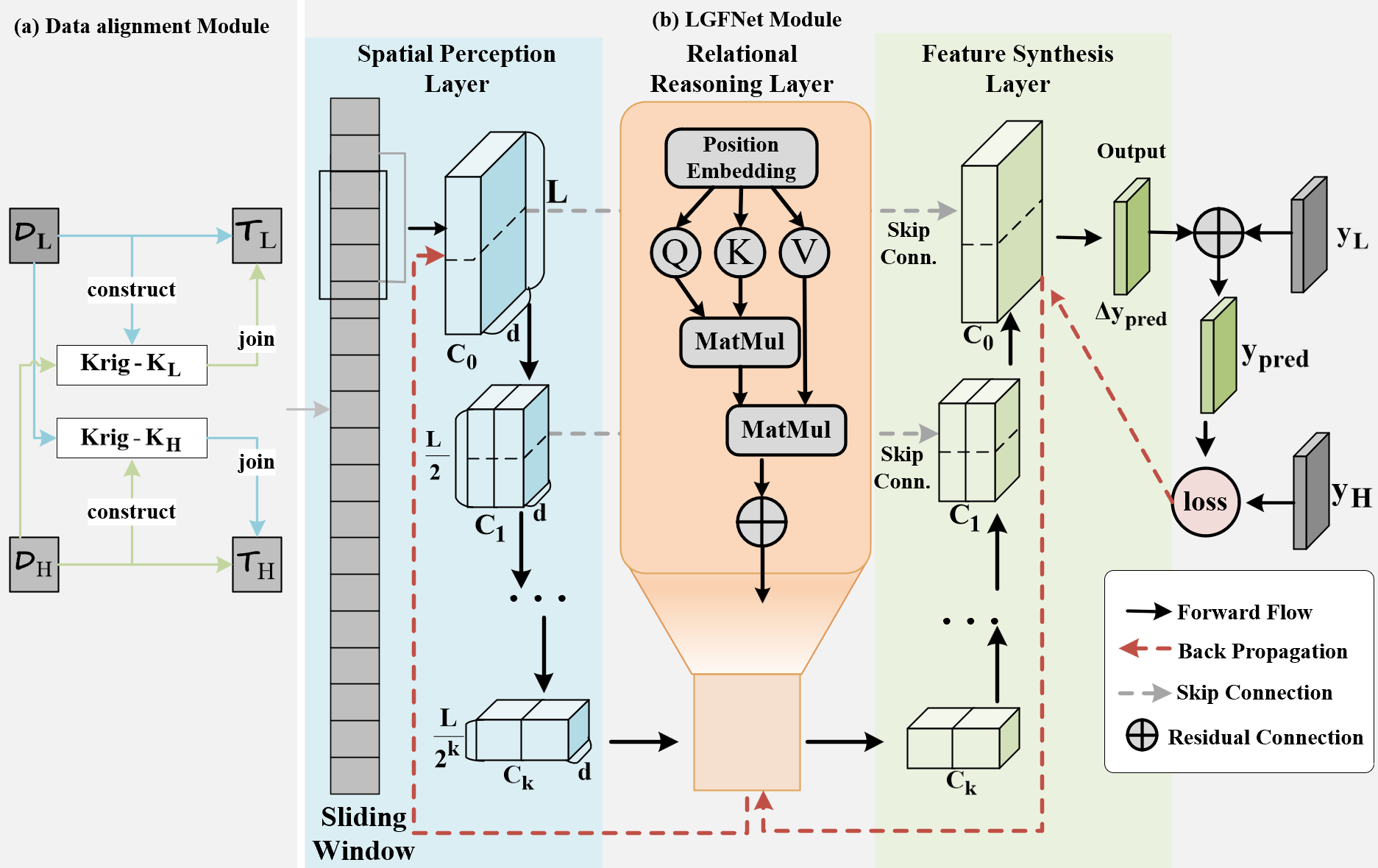} 
	\caption{The architecture of Local-Global Fusion Network (LGFNet) model. (a) data alignment module, (b) LGFNet module.}
	\label{FIG:LGFNet}
\end{figure*}

To effectively address the fusion of heterogeneous aerodynamic data, this paper proposes the Local-Global Fusion Network (LGFNet). As shown in Fig. \ref{FIG:LGFNet}, the model adopts an end-to-end pyramid architecture designed for multi-scale feature decomposition. The model is composed of a Data Alignment Module and a LGFNet Core Module. The core module integrates three specialized layers: the SPL establishes a sliding-window mechanism to reinforce the coverage and sequential continuity of local features ; the RRL introduces a self-attention mechanism to strengthen long-range dependencies by dynamically weighing global data relationships ; and the FSL reconstructs high-fidelity responses via skip connections to preserve fine-grained details. By synergizing these mechanisms, LGFNet provides a robust paradigm for capturing both sharp local discontinuities and macro-scale aerodynamic correlations.

Positing that the mapping from low-fidelity CFD simulations to high-fidelity measurements conceals a highly nonlinear "Fidelity Gap" \cite{Kennedy2000,Perdikaris2017}, we propose the FGDL strategy to explicitly capture this deviation. This strategy treats CFD data as a "low-frequency carrier", decoupling geometry-governed base flows from high-frequency nonlinear features such as shock waves.

In the context of Local-Global Fusion, FGDL serves as a critical learning paradigm that guides the network's focus. By reformulating the fusion task into a residual learning process, LGFNet is tasked with approximating the nonlinear discrepancy $\mathcal{R}_{\text{gap}}$, which represents the systematic bias between heterogeneous data sources. The fundamental relationship between the high-fidelity (HF) ground truth $\mathbf{y}_H$ and the low-fidelity (LF) estimate $\mathbf{y}_L$ is defined as:
\begin{equation}
\mathbf{y}_H(\mathbf{x}) = \mathbf{y}_L(\mathbf{x}) + \mathcal{R}_{\text{gap}}(\mathbf{x}, \mathbf{y}_L(\mathbf{x})),
\label{eq:fidelity_relation}
\end{equation}
where $\mathbf{x}$ denotes the aerodynamic state vector. Unlike standard residual learning that relies solely on $\mathbf{x}$, our implementation utilizes $\mathbf{y}_L(\mathbf{x})$ not only as the baseline carrier but also as an intrinsic input feature. This decoupling mechanism allows the SPL to perceive sharp gradients and physical discontinuities directly from the LF distribution, while the RRL captures the global topological shifts inherent in the discrepancy.

Within our proposed framework, LGFNet is employed to generate a predicted residual by observing the augmented feature vector $\mathbf{d} = [\mathbf{x}, \mathbf{y}_L]$. The prediction is formalized as:
\begin{equation}
\Delta \mathbf{y}_{\text{pred}} = \text{LGFNet}(\mathbf{x}, \mathbf{y}_L(\mathbf{x})).
\end{equation}

The optimization objective is to minimize the divergence between this predicted residual and the inherent system deviation. Mathematically, the ideal objective function $\mathcal{J}$ is formalized as:
\begin{equation}
\label{eq:objective_function}
\begin{split}
\mathcal{J} &= \mathbb{E}_{\mathbf{x} \sim \mathcal{D}} \left[ \left\| \Delta \mathbf{y}_{\text{pred}} - \mathcal{R}_{\text{gap}}(\mathbf{x}, \mathbf{y}_L(\mathbf{x})) \right\|_2^2 \right] \\
&= \mathbb{E}_{\mathbf{x} \sim \mathcal{D}} \left[ \left\| \text{LGFNet}(\mathbf{x}, \mathbf{y}_L(\mathbf{x})) - (\mathbf{y}_H(\mathbf{x}) - \mathbf{y}_L(\mathbf{x})) \right\|_2^2 \right]
\end{split}
\end{equation}
where $\mathbb{E}_{\mathbf{x} \sim \mathcal{D}}$ denotes the expectation over the data distribution $\mathcal{D}$ within the flight envelope, and $\|\cdot\|_2$ represents the $L_2$ norm used to quantify the error energy.

\subsection{Data alignment module}

We employ a Kriging interpolation model to achieve data alignment \cite{Jones2001}. It provides a method for unbiased estimation of the attribute value at an unknown point in a space using spatial partial information. For more details, refer to \cite{Han2016}.

For $\mathcal{D}_L$ and $\mathcal{D}_H$, we construct $K_L=\text{Kriging}(\mathcal{D}_L)$ and $K_H=\text{Kriging}(\mathcal{D}_H)$. We define the aerodynamic state vectors that exist in $\mathcal{D}_L$ but not in $\mathcal{D}_H$ as $S_1=\{\mathbf{x}|\mathbf{x} \in \mathcal{X}_L, \mathbf{x} \notin \mathcal{X}_H\}$, and similarly, $S_2=\{\mathbf{x}|\mathbf{x} \in \mathcal{X}_H, \mathbf{x} \notin \mathcal{X}_L\}$, where $\mathcal{X}_L=\{x_i\}_{i=1}^{N_L}$ and $\mathcal{X}_H=\{x_j\}_{j=1}^{N_H}$. Then, we perform cross-interpolation:
\begin{equation}
\begin{aligned}
y_H(\mathbf{x}) &= K_H(\mathbf{x}) \\
y_L(\mathbf{x}) &= K_L(\mathbf{x})
\end{aligned}
\end{equation}

It is worth noting that the Kriging interpolation model has poor extrapolation ability and mostly performs interpolation within the interpolation range \cite{XU2022,XU2024}. Therefore, after constructing the Kriging interpolation models, the newly interpolated aerodynamic state vectors should not exceed the numerical range of the original datasets. Consequently, the intersection of the numerical ranges of $\mathcal{X}_L$ and $\mathcal{X}_H$ from $\mathcal{D}_L$ and $\mathcal{D}_H$ is selected as the aerodynamic state range $\mathcal{X}_L \cap \mathcal{X}_H$ for the aligned datasets $\mathcal{T}_L$ and $\mathcal{T}_H$. $\mathcal{T}_L$ and $\mathcal{T}_H$ are used as the input and label of LGFNet module.

\subsection{LGFNet module}

\subsubsection {Spatial perception layer}

The SPL is specifically designed to handle the multi-scale characteristics of aerodynamic flow fields by reinforcing the coverage and sequential continuity of local features.

\begin{figure*}
	\centering
	\includegraphics[width=0.8\textwidth]{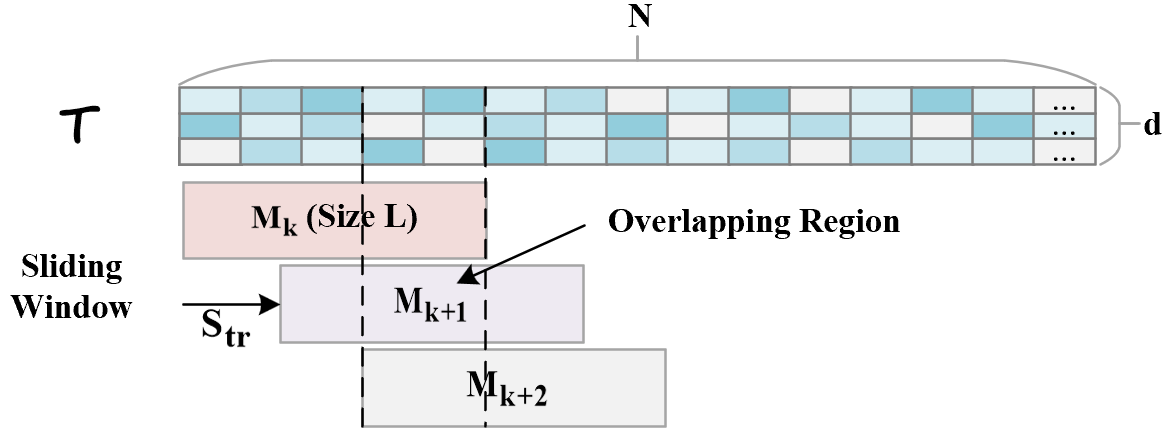} 
	\caption{Sliding Window (SW) mechanism.}
	\label{FIG:SW}
\end{figure*}

Let the aligned global dataset be denoted as $\mathcal{T} \in \mathbb{R}^{N \times d}$, where $N$ is the total number of samples and $d$ is the dimension of the unified aerodynamic vector $\mathbf{d}$.The process begins with Sliding Window (SW) mechanism. Unlike standard batch processing, SW mechanism preserves the sequential continuity required to detect local abrupt changes (e.g., shock waves).The SW mechanism is shown in the Fig. \ref{FIG:SW}.  We define the sliding window transformation as a mapping that generates a set of overlapping local blocks $\mathcal{M} = \{ \mathbf{M}_k \}_{k=1}^{N_{str}}$:
\begin{equation}
\mathbf{M}_k = \mathcal{T}[s_k : s_k + L, :] \in \mathbb{R}^{L \times d}, \quad s_k = (k-1) \cdot S_{tr}
\end{equation}
where $L$ denotes the window size, $S_{tr}$ is the stride, and $s_k$ represents the starting index of the $k$-th block. This formulation ensures that the local gradient information within the window $L$ is preserved, serving as a physics-informed data augmentation strategy.

To process these continuous local sequences effectively, we construct a hierarchical encoder consisting of $K$ stages. 
Let the batch size be $b$ and the initial number of channels be $C_0$. The input batch tensor to the encoder can be represented as $\mathbf{E}_0 \in \mathbb{R}^{b \times C_0 \times L \times d}$.

Each encoder block serves a dual purpose: increasing feature abstraction capability through channel expansion and reducing computational complexity through sequence length compression. 
Let $\mathbf{E}_{i-1}$ denote the input feature tensor for the $i$-th stage with channel dimension $C_{i-1}$ and sequence length $L_{i-1}$. The transformation process in the $i$-th stage is formalized as follows:

\begin{equation}
\left\{
\begin{aligned}
\mathbf{E}_i^{(1)} &= \text{ReLU}\left(\text{BN}\left(\text{Conv}_{2D}(\mathbf{E}_{i-1})\right)\right) \\
\mathbf{E}_i^{(2)} &= \text{ReLU}\left(\text{BN}\left(\text{Conv}_{2D}(\mathbf{E}_i^{(1)})\right)\right) \\
\mathbf{E}_i &= \text{MaxPool2D}\left(\mathbf{E}_i^{(2)}, (2,1)\right)
\end{aligned}
\right.
\end{equation}

Through this hierarchical stacking, the tensor dimensions evolve dynamically: the channel dimension doubles $C_{i} = 2C_{i-1}$ to capture richer high-level semantic features of the flow field, while the sequence length halves $L_{i} = L_{i-1}/2$ via the anisotropic pooling. 

Critically, the anisotropic kernel selectively compresses the spatial dimension $L$ while keeping the aerodynamic feature dimension $d$ invariant. This design ensures that the intrinsic topological correlations among coupled aerodynamic variables (e.g., the relationship between $Ma$ and $C_p$) are strictly preserved throughout the deep feature perception process. 

Consequently, the final output of the SPL is a high-dimensional feature tensor $\mathbf{E}_k \in \mathbb{R}^{b \times C_k \times L_k \times d}$, encoding dense local spatial patterns.

\subsubsection{Relational reasoning layer}

Although the SPL excels at capturing local details, its convolutional nature inherently prioritizes neighborhood correlations, often overlooking the dynamic global couplings prevalent in aerodynamic flows—such as the correlation between leading-edge features and trailing-edge separation. Conventional fixed-weight kernels struggle to model these input-dependent, long-range associations across diverse flight states.

To overcome this, the RRL introduces Multi-Head Self-Attention (MHSA) to capture global topological structures. By providing a basic perceptual domain through downsampling, the MHSA module further enhances long-range dependencies by dynamically weighing data relationships. Furthermore, since the standard softmax normalization self-attention mechanism is essentially a low-pass filter, it naturally attenuates high-frequency signal components and smooths out the high-frequency interpolation noise introduced during the Kriging data alignment stage. This mechanism is essential for understanding complex global flow interactions, allowing the model to adaptively bridge distant but physically coupled aerodynamic states within a unified Local-Global Fusion framework.

First, the local feature tensor $\mathbf{E}_k$ is reshaped into a unified sequence matrix $\mathcal{T} \in \mathbb{R}^{b \times L_B \times d}$, where $L_B = C_k \times L_k$. 
This transformation flattens the channel and spatial dimensions, allowing the model to treat all extracted local features as a global token sequence for holistic reasoning. 
The MHSA mechanism with $h$ heads is mathematically formulated as:

\begin{equation}
\mathcal{T}_{pos} = \mathcal{T} + \text{PE}(\mathcal{T})
\end{equation}
\begin{equation}
\mathbf{Q}_i = \mathcal{T}_{pos}\mathbf{W}_Q, \quad \mathbf{K}_i = \mathcal{T}_{pos}\mathbf{W}_K, \quad \mathbf{V}_i = \mathcal{T}_{pos}\mathbf{W}_V
\end{equation}
\begin{equation}
\text{head}_i = \text{softmax}\left(\frac{\mathbf{Q}_i \mathbf{K}_i^T}{\sqrt{d_k}}\right)\mathbf{V}_i
\end{equation}
\begin{equation}
\mathcal{T}_{att} = \text{Concat}(\text{head}_1, \dots, \text{head}_h)\mathbf{W}_O
\end{equation}

$\text{PE}(\cdot)$ denotes the sine and cosine positional encoding to retain sequence order information. $\mathbf{W}_Q, \mathbf{W}_K, \mathbf{W}_V, \mathbf{W}_O$ are learnable projection matrices.

\textbf{Query ($\mathbf{Q}$):} Represents the target aerodynamic states for which the model seeks to establish correlations. It acts as a "probe" to inquire about the influence of other flow regions. \textbf{Key ($\mathbf{K}$):} Represents the global characteristic indices of the aerodynamic data. It encodes the identity of flow patterns (e.g., presence of a shock wave or vortex) at different positions. \textbf{Value ($\mathbf{V}$):} Stores the actual information content of the aerodynamic forces/moments. It contains the physical feature representations to be aggregated.

The attention score matrix, computed via $\mathbf{Q}\mathbf{K}^T$, quantifies the intensity of physical concern or correlation between different aerodynamic states. 
High attention weights indicate a strong physical coupling (e.g., how the upstream Angle of Attack stored in $\mathbf{K}$ affects the downstream pressure distribution queried by $\mathbf{Q}$), effectively bridging the "fidelity gap" by leveraging global context to calibrate local predictions.

Finally, the globally refined features are superimposed onto the original local features via a residual connection to yield the input for the integration layer:
\begin{equation}
\mathbf{E}'_k = \text{Reshape}(\mathcal{T}_{att}) + \mathbf{E}_k
\end{equation}
This design ensures that the model retains precise local details from the SPL while benefiting from the global aerodynamic context captured by the reasoning mechanism.

\subsubsection{Feature synthesis layer}
The FSL functions as a symmetric decoder to reconstruct high-resolution aerodynamic responses by organically synthesizing the globally reasoned features with the reinforced local spatial details. It progressively restores the sequence length to the original physical window size $L$.

To ensure that the macro-scale topological structures captured by the attention mechanism do not compromise the resolution of sharp local features (such as shock waves), the synthesis process relies on skip connections. These connections bridge the ``semantic gap'' between the encoder and decoder, facilitating a multi-scale fusion of features. In each of the $K$ decoding stages, the upsampled global features are concatenated with the corresponding local features $\mathbf{E}_i$ preserved during the perception stage. The transformation in the $i$-th stage is formalized as follows:

\begin{equation}
\left\{
\begin{aligned}
\mathbf{D}_i^{cat} &= \text{Concat}\left(\text{Up}\left(\mathbf{D}_{i+1}\right), \mathbf{E}_i\right) \\
\mathbf{D}_i^{(1)} &= \text{ReLU}\left(\text{BN}\left(\text{Conv}_{2D}(\mathbf{D}_i^{cat})\right)\right) \\
\mathbf{D}_i &= \text{ReLU}\left(\text{BN}\left(\text{Conv}_{2D}(\mathbf{D}_i^{(1)})\right)\right)
\end{aligned}
\right.
\end{equation}

where $\text{Up}(\cdot)$ denotes the bilinear upsampling operation used to restore the temporal resolution, and $\mathbf{D}_i^{cat}$ represents the fused feature map at the $i$-th decoding
stage. This hierarchical synthesis ensures that the refined features effectively recover complex, non-linear aerodynamic patterns across the entire flight envelope.

Finally, a $1 \times 1$ convolutional layer is employed as the output projection head. It maps the synthesized local-global feature representation into the target response dimension, yielding the predicted aerodynamic residual:
\begin{equation}
\Delta \mathbf{y}_{\text{pred}} = \text{Conv}_{1 \times 1}(\mathbf{D}_1) \in \mathbb{R}^{b \times 1 \times L \times d_y} 
\end{equation}

By integrating high-frequency local corrections with low-frequency global trends, the FSL achieves a high-fidelity reconstruction of multi-source aerodynamic data.

\subsubsection{Fidelity gap delta learning strategy}

The optimization of LGFNet follows the FGDL strategy, which serves as the training paradigm to consolidate the results of Local-Global Fusion. Rather than directly predicting high-fidelity aerodynamic responses, the loss function is designed to supervise the model's ability to reconstruct the nonlinear discrepancy $\mathcal{R}_{\text{gap}}$ between heterogeneous sources.

We premise that while low-fidelity CFD data ($\mathbf{y}_L$) captures fundamental physical trends, it lacks the high-frequency local precision found in experimental truths ($\mathbf{y}_H$). To bridge this gap, the model is trained to minimize the Mean Squared Error (MSE) between the predicted residual, synthesized from local-global features, and the ground-truth deviation. Following the augmented input logic, the loss function $\mathcal{L}$ is formalized as:

\begin{equation}
    \label{eq:loss_function}
    \begin{split}
        \mathcal{L} &= \frac{1}{b} \sum_{k=1}^{b} \left\| \Delta \mathbf{y}_{\text{pred}}^{(k)} - \Delta \mathbf{y}_{\text{true}}^{(k)} \right\|_2^2 \\
        &= \frac{1}{b} \sum_{k=1}^{b} \left\| \text{LGFNet}(\mathbf{x}_k, \mathbf{y}_{L,k}) - (\mathbf{y}_{H,k} - \mathbf{y}_{L,k}) \right\|_2^2
    \end{split}
\end{equation}
where $b$ denotes the batch size, and $\Delta \mathbf{y}_{\text{pred}}^{(k)}$ is the output of the FSL for the $k$-th sample, generated by observing both the state $\mathbf{x}_k$ and the low-fidelity carrier $\mathbf{y}_{L,k}$.

During the inference phase, the final high-fidelity prediction $\hat{\mathbf{y}}$ is obtained by superimposing the synthesized residual onto the original low-fidelity carrier:
\begin{equation}
    \hat{\mathbf{y}} = \mathbf{y}_L + \Delta \mathbf{y}_{\text{pred}}
\end{equation}
The FGDL strategy allows LGFNet to effectively inherit the broad physical trends of CFD data while achieving superior numerical accuracy through the fusion of precisely reasoned global contexts and reinforced local details.

To provide a clear and comprehensive overview of the proposed methodology, the complete training process of LGFNet, governed by the FGDL strategy, is summarized in Algorithm \ref{alg:LGFNet_training}.

\begin{algorithm}[H]
    \caption{Training Process of LGFNet}
    \label{alg:LGFNet_training}
    \begin{algorithmic}[1]
        \Require Aligned aerodynamic states $X$, Low-fidelity responses $Y_L$, High-fidelity responses $Y_H$, Max epochs $N_{epoch}$.
        \Ensure Optimal LGFNet parameters $\theta = \{\theta_{SPL}, \theta_{RRL}, \theta_{FSL}\}$.
        
        \State $\Delta Y_{true} \leftarrow Y_H - Y_L$ \Comment{Decouple the ground-truth fidelity gap}
        \State $\mathcal{D}_{train} \leftarrow \text{SlidingWindow}(X, Y_L, \Delta Y_{true})$ \Comment{Generate sequential local blocks}
        
        \For{$epoch = 1$ \textbf{to} $N_{epoch}$}
            \For{\textbf{each} batch $(x, y_L, \Delta y_{true}) \in \mathcal{D}_{train}$}
                
                \State \textbf{/* 1. SPL */}
                \State $E_k, \{E_{skip}\} \leftarrow \text{SPL}([x, y_L]; \theta_{SPL})$ \Comment{Extract local multi-scale discontinuities}
                
                \State \textbf{/* 2. RRL */}
                \State $E'_k \leftarrow E_k + \text{MHSA}(\text{PE}(E_k); \theta_{RRL})$ \Comment{Capture global topological dependencies}
                
                \State \textbf{/* 3. FSL */}
                \State $\Delta y_{pred} \leftarrow \text{FSL}(E'_k, \{E_{skip}\}; \theta_{FSL})$ \Comment{Reconstruct residuals via skip connections}
                
                \State \textbf{/* 4. FGDL (Optimization) */}
                \State $\mathcal{L} \leftarrow ||\Delta y_{pred} - \Delta y_{true}||_2^2$ \Comment{Minimize non-linear discrepancy}
                \State $\theta \leftarrow \text{Adam}(\theta, \nabla_\theta \mathcal{L})$ \Comment{Update network parameters}
                
            \EndFor
        \EndFor
        \State \Return $\theta$
        \end{algorithmic}
        \end{algorithm}

\section{Experiments}
\subsection{Multi-source aerodynamic datasets}
According to the physical characteristics of aerodynamic data, we formulate the fusion task into the following two scenarios:

\begin{itemize}
    \item \textbf{Scenario 1: Distribution fusion.} This scenario targets the pressure coefficient $C_p$ distribution on airfoil surfaces. Here, the input feature vector $\mathbf{x}$ comprises flow conditions (e.g., Mach number $Ma$, Angle of Attack $\alpha$) and spatial geometric coordinates. The target $\mathbf{y}$ is the corresponding scalar pressure value. The model is required to capture the spatial evolution laws of the flow field along the chord length.
    \item \textbf{Scenario 2: Coefficient fusion.} This scenario targets the aerodynamic coefficients ($C_x, C_y, C_z$) of aircraft under complex operating conditions. Here, the input $\mathbf{x}$ consists of high-dimensional aerodynamic state variables (e.g., $Ma, \alpha, \beta$, etc.), and the target $\mathbf{y}$ is the vector of aerodynamic force/moment coefficients. The model needs to establish a non-linear mapping from the high-dimensional state space to the aerodynamic response.
\end{itemize}

To evaluate the proposed framework, two datasets representing distinct fusion scenarios are employed in experiments: the transonic RAE2822 airfoil for Scenario 1 (Distribution Fusion), which focuses on surface pressure distributions, and a CARDC aircraft model for Scenario 2 (Coefficient Fusion), which targets concentrated force coefficients in high-dimensional state spaces.

\subsubsection{RAE2822 airfoil dataset}
The first case considers the RAE2822 airfoil. In this study, the task is to predict and fuse the pressure coefficient ($C_p$) distribution along the normalized chord length ($x/c$) obtained from CFD simulations and wind tunnel tests.

\textbf{Low-fidelity data source:} the low-fidelity data are obtained through CFD simulations. These simulations are conducted using the commercial solver ANSYS Fluent, solving the Reynolds-Averaged Navier-Stokes (RANS) equations. For each operating case, the CFD simulation generates 422 data points characterizing the pressure distribution on the upper and lower airfoil surfaces.

\textbf{High-fidelity data source:} the high-fidelity data are sourced from the wind tunnel test measurements (Wind) provided by Cook et al. \cite{cook1979} in the AGARD advisory report. The test datasets for Case 1, Case 2, and Case 3 contain 103, 102, and 100 discrete measurement points, respectively. These test data serve as the ground truth for validating the fusion accuracy.

To verify the robustness of the model under varying flow conditions, three distinct operating cases are selected, covering subsonic and transonic regimes with different angles of attack. The specific inflow conditions for these test cases are detailed in Table \ref{tab:rae_conditions}.

\begin{figure*}[htbp]
    \centering
    \begin{subfigure}[b]{0.32\textwidth}
        \centering
        \includegraphics[width=\linewidth]{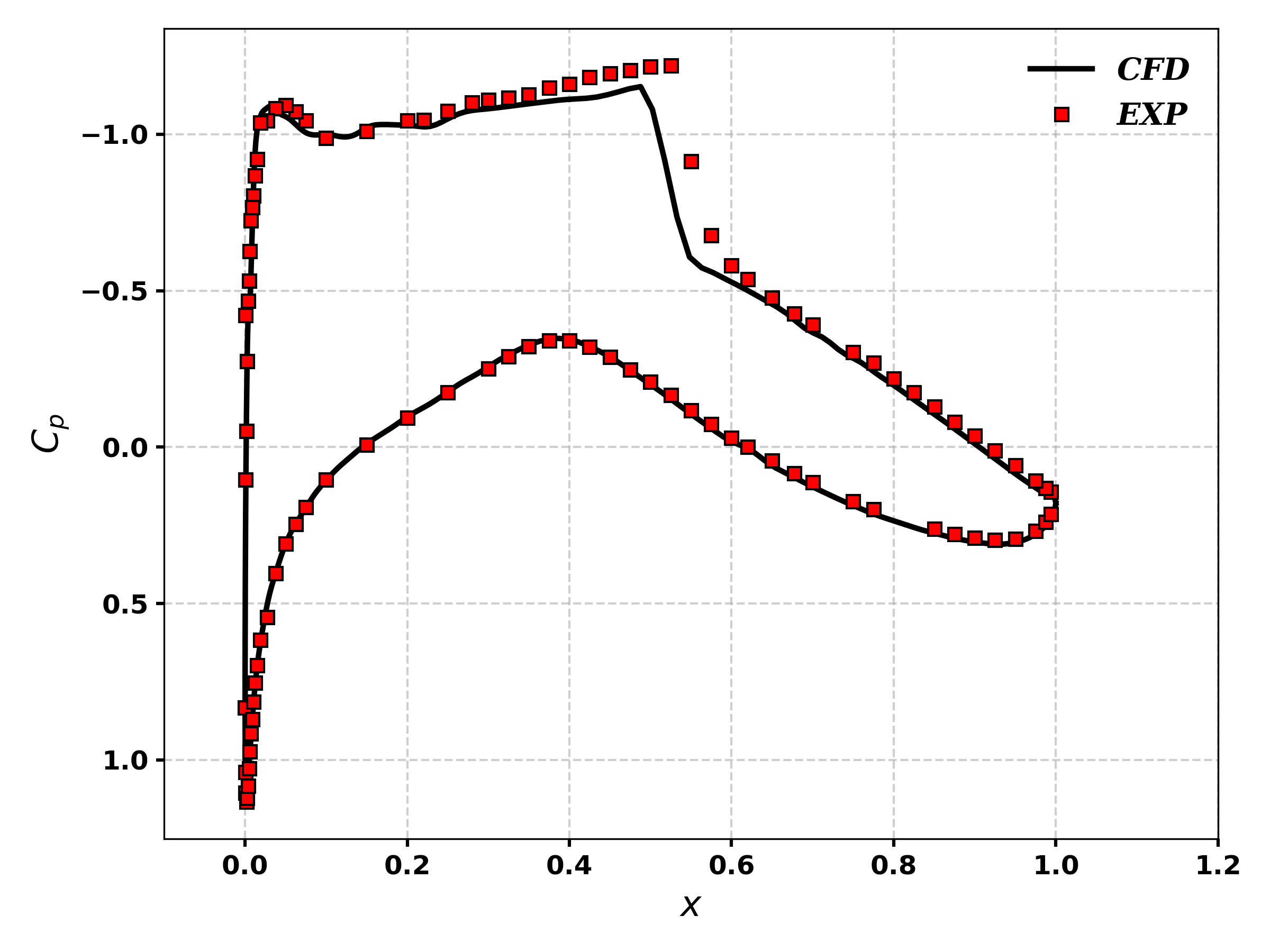} 
        \caption{Case 1}
        \label{fig:case1}
    \end{subfigure}
    \hfill 
    \begin{subfigure}[b]{0.32\textwidth}
        \centering
        \includegraphics[width=\linewidth]{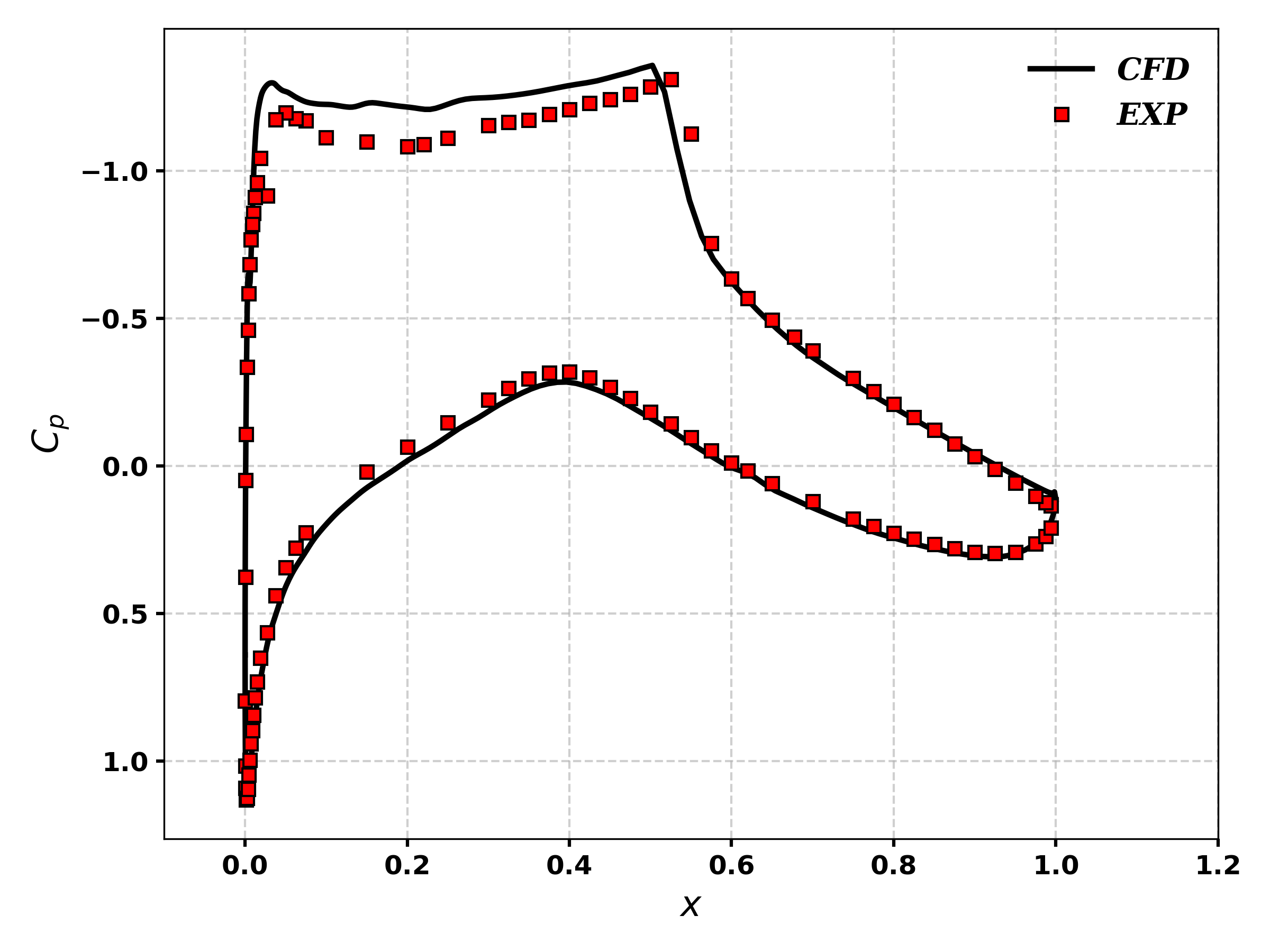} 
        \caption{Case 2}
        \label{fig:case2}
    \end{subfigure}
    \hfill
    \begin{subfigure}[b]{0.32\textwidth}
        \centering
        \includegraphics[width=\linewidth]{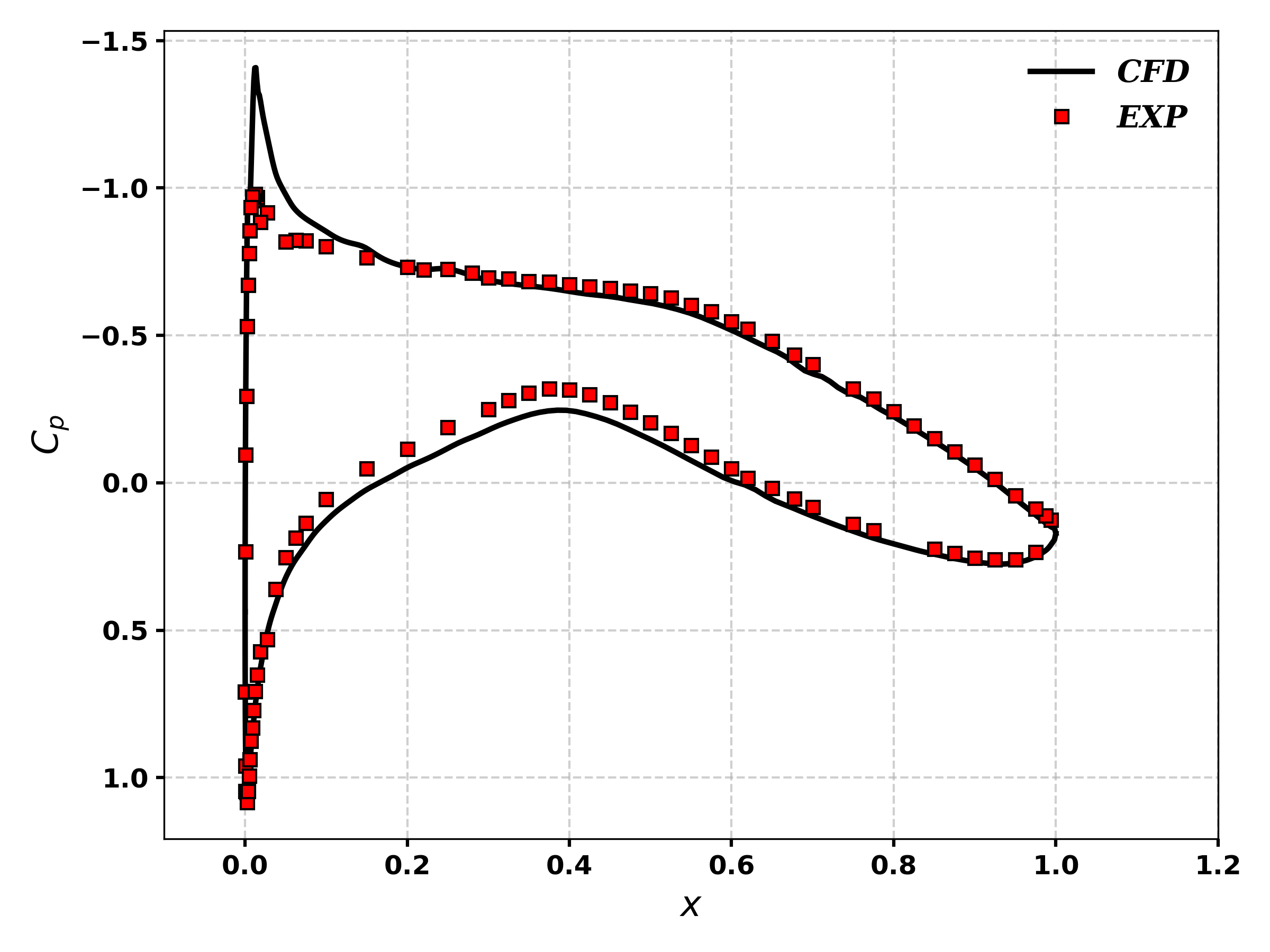} 
        \caption{Case 3}
        \label{fig:case3}
    \end{subfigure}
    
    \caption{Comparison between CFD simulations and Wind data.}
    \label{FIG:data_comparison}
\end{figure*}

The preliminary comparison between the low-fidelity CFD simulations and high-fidelity wind tunnel test data is illustrated in Fig. \ref{FIG:data_comparison}. It can be observed that:
\begin{itemize}
    \item \textbf{Transonic discrepancies:} in Case 1 and Case 2, which operate in the transonic regime, a significant "fidelity gap" exists. The CFD simulations fail to accurately capture the sharp pressure rise associated with the shock wave on the upper surface ($x/c \approx 0.5$). The experimental data exhibit a much steeper discontinuity compared to the smeared gradients predicted by the RANS-based CFD.
    \item \textbf{Systematic biases:} in the subsonic Case 3, although the shock wave is absent, systematic deviations in the peak suction and trailing-edge pressure recovery remain evident between the two sources. 
\end{itemize}

\begin{table}[width=.9\linewidth,cols=4,pos=ht]
\caption{The inflow conditions for the RAE2822 airfoil test cases.}\label{tab:rae_conditions}
\begin{tabular*}{\tblwidth}{@{} LLLL@{} }
\toprule
Case & Mach ($Ma$) & Reynolds ($Re$) & AoA ($\alpha$) \\
\midrule
Case 1 & 0.729 & $6.5 \times 10^6$ & $2.31^\circ$ \\
Case 2 & 0.730 & $6.5 \times 10^6$ & $3.19^\circ$ \\
Case 3 & 0.600 & $6.3 \times 10^6$ & $2.57^\circ$ \\
\bottomrule
\end{tabular*}
\end{table}

\subsubsection{CARDC aircraft dataset}
The second case involves a real-world engineering application using data from a specific aircraft model provided by the China Aerodynamics Research and Development Center (CARDC). Unlike the airfoil case, this dataset represents a concentrated force fusion problem within a high-dimensional state space.

\textbf{Data sources:} the low-fidelity dataset consists of numerical calculation results obtained from CFD simulations, while the high-fidelity dataset is acquired from actual flight tests. Specifically, the CFD dataset comprises 38,116 data samples, whereas the flight test (FLI) dataset contains 16,983 samples. Due to the high cost and risk associated with flight tests, the high-fidelity samples are significantly sparser relative to the potential flight envelope than the CFD samples.

\textbf{Input and output definition:} both data sources share a consistent feature space. The input vector $x$ consists of six aerodynamic state variables: $\mathbf{x}=(Ma, \alpha, \phi, D_x, D_y, D_z)$, where $Ma$ is Mach number, $\alpha$ is angle of attack (AoA), $\phi$ is sideslip angle, and $D_x, D_y, D_z$ are control surface deflections. The output vector $\mathbf{y}$ corresponds to the three-dimensional aerodynamic force coefficients: $\mathbf{y}=(C_x, C_y, C_z)$, where $C_x$ is axial force coefficient, $C_y$ is normal force coefficient, and $C_z$ is lateral force coefficient. The fusion task is performed individually for each force coefficient based on the input aerodynamic states.

\subsection{Experimental setup}
\subsubsection{Network Configuration}
The proposed LGFNet framework is implemented using the PyTorch deep learning library.
LGFNet employs a unified architecture for both fusion scenarios.
In sliding window mechanism (SW) of the encoder, the window length is set to $L=112$ with a stride of $S=14$.
The network width is adaptively configured to address the specific challenges of each task:
\begin{itemize}
    \item \textbf{Scenario 1 (Pressure distribution):} to capture fine-grained local flow features like shock waves, the channel dimensions are set to $[32, 64, 128, 256, 512]$ with a batch size of 64.
    \item \textbf{Scenario 2 (Aerodynamic coefficients):} to prevent overfitting on the high-dimensional manifold, a lightweight configuration with channels $[8, 16, 32, 64, 128]$ and a batch size of 64 is adopted.
\end{itemize}
In the Relational Reasoning Layer, a multi-head self-attention module (1 head, dropout rate 0.1) is integrated to capture global dependencies.

\subsubsection{Implementation details}
Network parameters are optimized using the Adam optimizer with an initial learning rate of $5 \times 10^{-4}$.An adaptive learning rate scheduler (ReduceLROnPlateau) is employed, decaying the learning rate by a factor of 0.5 if the training loss stagnates for 15 consecutive epochs.

\subsubsection{Training and inference strategy}
To ensure rigorous validation suitable for the distinct characteristics of each dataset, specific training and inference workflows were designed. A detailed comparison of the strategies is summarized in Table \ref{tab:workflow_strategies}.

\begin{table*}[htbp]
    \centering
    \caption{\centering Comparison of training and inference workflows.}
    \label{tab:workflow_strategies}
    \renewcommand{\arraystretch}{1.4}
    \begin{tabularx}{\textwidth}{@{} l X X @{}}
    \toprule
    Workflow Phase & Scenario 1: RAE2822 Airfoil & Scenario 2: CARDC Aircraft \\
    \midrule
    Primary Challenge & 
    Scarcity of discrete operating cases & 
    Continuity of high-dimensional trajectories \\
    
    Data Partitioning & 
    Case-wise rotating split via ``Leave-Half-Out'' strategy (training : testing = 2.5 : 0.5) & 
    Block-wise cyclical split via Mach number jumps (training : testing = 4 : 1) \\
    
    Training Strategy & 
    Hyperparameter tuning via 5-fold cross-validation, followed by full-set retraining & 
    Data augmentation via sliding window mechanism, followed by standard network training\\
    
    Evaluation Metric & 
    \multicolumn{2}{c}{Quantitative assessment on the sealed test set using RMSE, MAE, and $R^2$} \\
    
    Final Inference & 
    \multicolumn{2}{c}{Comprehensive database generation via full-sequence inference} \\
    \bottomrule
    \end{tabularx}
    \end{table*}

Although applying Kriging alignment prior to the train/test split poses a theoretical risk of data leakage, its impact is mathematically negligible. First, Kriging serves strictly as a distance-based spatial resampling tool for coordinate alignment rather than a feature extractor \cite{Han2016, Cressie1993}. Second, due to the exponential decay of spatial autocorrelation, the interpolated values within the training grids are dominated exclusively by their adjacent training samples \cite{Rasmussen2006}. Consequently, the mathematical influence of the separated test set on the training data approaches zero, ensuring that the true non-linear fidelity gap is exclusively learned by LGFNet within a strictly isolated training environment.

\subsubsection{Uncertainty Quantification}
\label{sec:uncertainty}

The predictive uncertainty $U$ is quantified as the expected width of the $1-\alpha$ confidence interval over the test set:
\begin{equation}
\label{eq:uncertainty}
\begin{aligned}
U &= \frac{1}{N_{test}} \sum_{i=1}^{N_{test}} \left(Upper_{i} - Lower_{i}\right) \\
  &= \frac{2 z_{1-\alpha/2}}{N_{test}} \sum_{i=1}^{N_{test}} \sigma(\mathbf{x}_i)
\end{aligned}
\end{equation}
where $z_{1-\alpha/2}$ denotes the standard normal quantile, and $\sigma(\mathbf{x}_i)$ is the predictive standard deviation. 

\paragraph{Scenario 1: Exact GPR} 
Using the full training set $\mathbf{X}$ of size $N$, the exact predictive variance is formulated as:
\begin{equation}
\label{eq:sigma_exact}
\sigma_N^2(\mathbf{x}_i) = k(\mathbf{x}_i, \mathbf{x}_i) - \mathbf{k}_N^\top(\mathbf{x}_i) \left( \mathbf{K}_{NN} + \sigma_n^2 \mathbf{I} \right)^{-1} \mathbf{k}_N(\mathbf{x}_i)
\end{equation}
where $\mathbf{K}_{NN} = k(\mathbf{X}, \mathbf{X})$, $\mathbf{k}_N(\mathbf{x}_i) = k(\mathbf{X}, \mathbf{x}_i)$, and $\sigma_n^2$ is the noise variance.

\paragraph{Scenario 2: Sparse Approximation via FIC} 
An active subset $\mathcal{S} \subset \mathbf{X}$ of size $M$ ($M \ll N$) is selected by maximizing the marginal log-likelihood:
\[
\mathcal{S} = \underset{\mathcal{S} \subset \mathbf{X}, |\mathcal{S}| = M}{\arg\max} \log p(\mathbf{y}_{\mathcal{S}} \mid \mathbf{X}_{\mathcal{S}})
\]
Under the fully independent conditional (FIC) approximation, the predictive variance is calculated as:
\begin{equation}
    \label{eq:sigma_approx}
    \begin{split}
        \sigma_M^2(\mathbf{x}_i) &= k(\mathbf{x}_i, \mathbf{x}_i) - \mathbf{k}_M^\top(\mathbf{x}_i) \Big[ \mathbf{K}_{MM}^{-1} \\
        &\quad - \left( \mathbf{K}_{MM} + \mathbf{K}_{MN} \mathbf{\Lambda}^{-1} \mathbf{K}_{NM} \right)^{-1} \Big] \mathbf{k}_M(\mathbf{x}_i)
    \end{split}
    \end{equation}
where $\mathbf{K}_{MM} = k(\mathbf{X}_{\mathcal{S}}, \mathbf{X}_{\mathcal{S}})$ and $\mathbf{k}_M(\mathbf{x}_i) = k(\mathbf{X}_{\mathcal{S}}, \mathbf{x}_i)$. The diagonal matrix $\mathbf{\Lambda}$ compensates for the variance approximation error, defined as:
\[
\mathbf{\Lambda} = \text{diag} \left( \mathbf{K}_{NN} - \mathbf{K}_{MN}\mathbf{K}_{MM}^{-1}\mathbf{K}_{NM} \right) + \sigma_n^2 \mathbf{I}
\]
where $\mathbf{K}_{MN}$ is the cross-covariance matrix between $\mathcal{S}$ and $\mathbf{X}$.

\subsection{Comparison experiments}
\subsubsection{Results on RAE2822 airfoil dataset}

In this subsection, the proposed LGFNet framework is validated on the RAE2822 airfoil dataset across three distinct operating conditions (Case 1, Case 2, and Case 3). The prediction performance is compared with five baseline algorithms: Hierarchical Kriging (HK) \cite{Han2012}, DNN \cite{2022_Li}, MSFM \cite{Hu2025}, XGBoost GF is \cite{Lin2024} and ArGEnT is \cite{Chen2026ArGEnT}. The high-fidelity wind tunnel experimental data (WIND) serves as the ground truth. The quantitative metrics including Root Mean Square Error (RMSE), Mean Absolute Error (MAE), Coefficient of Determination ($R^2$), Training Time, and Uncertainty are summarized in Table \ref{tab:rae_results}.

\begin{figure*}[htbp]
    \centering
    \begin{subfigure}[b]{0.32\textwidth}
        \centering
        \includegraphics[width=\linewidth]{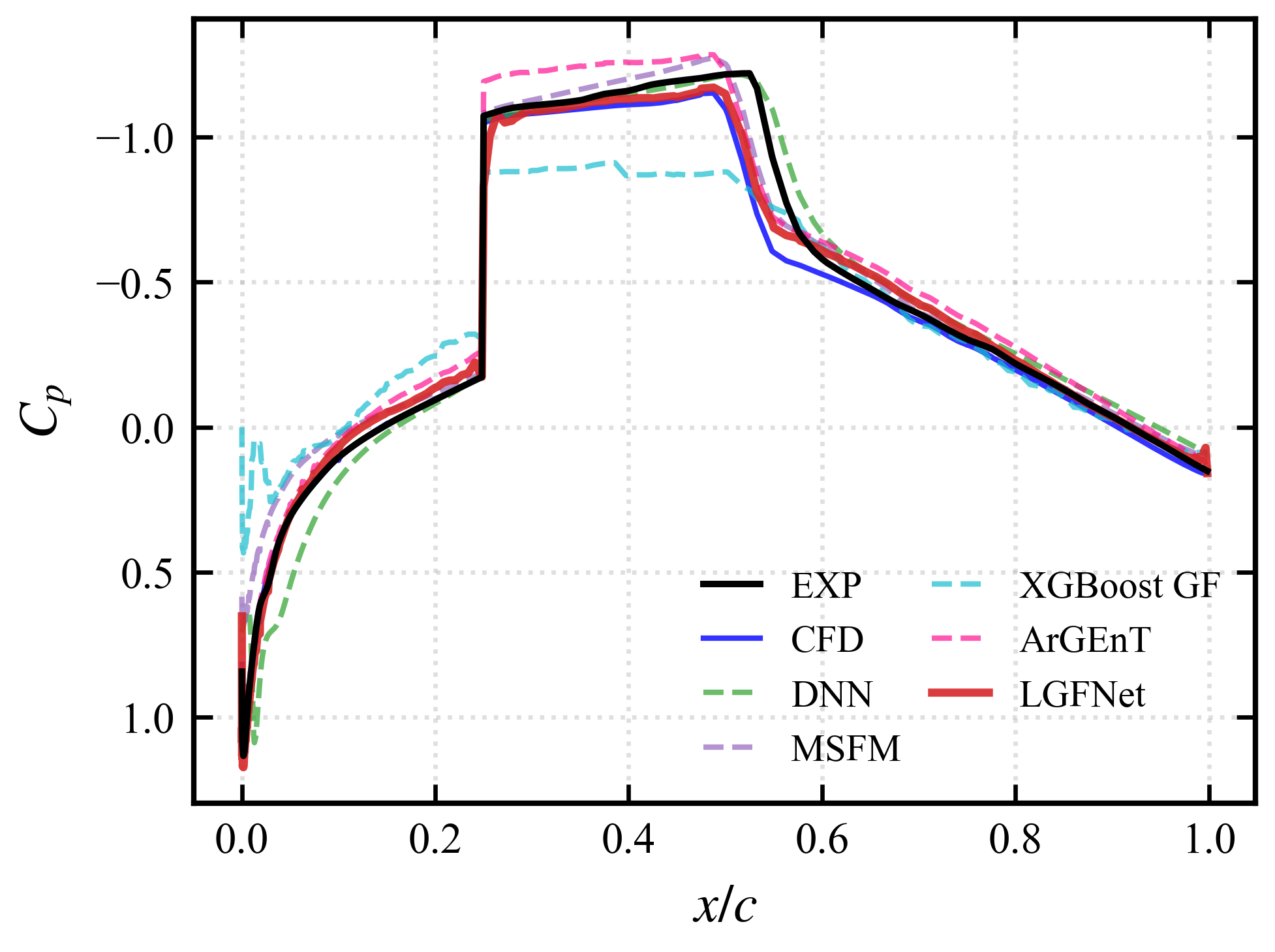} 
        \caption{Case 1: Test Dateset}
        \label{fig:case1_test}
    \end{subfigure}
    \hfill 
    \begin{subfigure}[b]{0.32\textwidth}
        \centering
        \includegraphics[width=\linewidth]{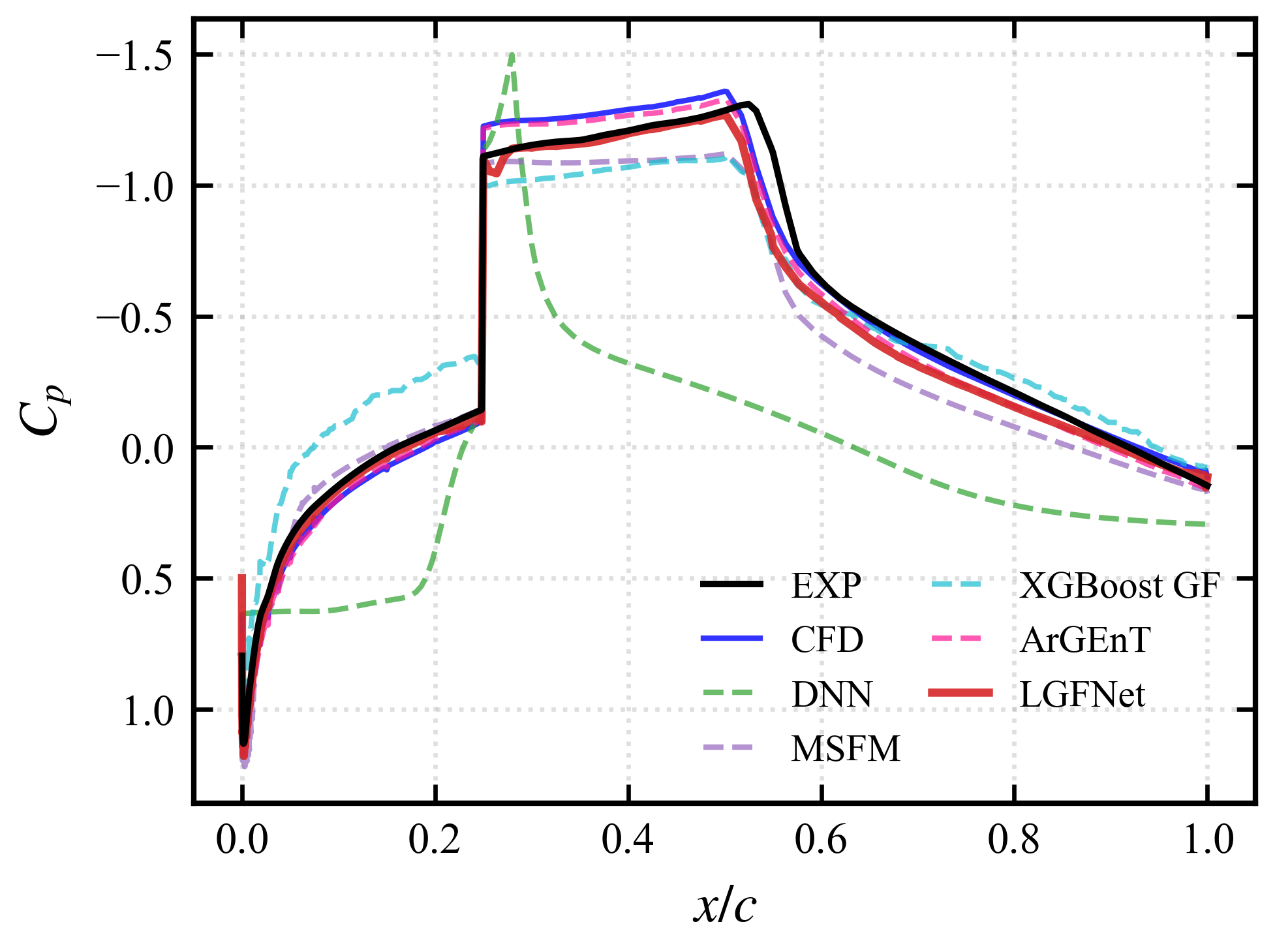} 
        \caption{Case 2: Test Dateset}
        \label{fig:case2_test}
    \end{subfigure}
    \hfill
    \begin{subfigure}[b]{0.32\textwidth}
        \centering
        \includegraphics[width=\linewidth]{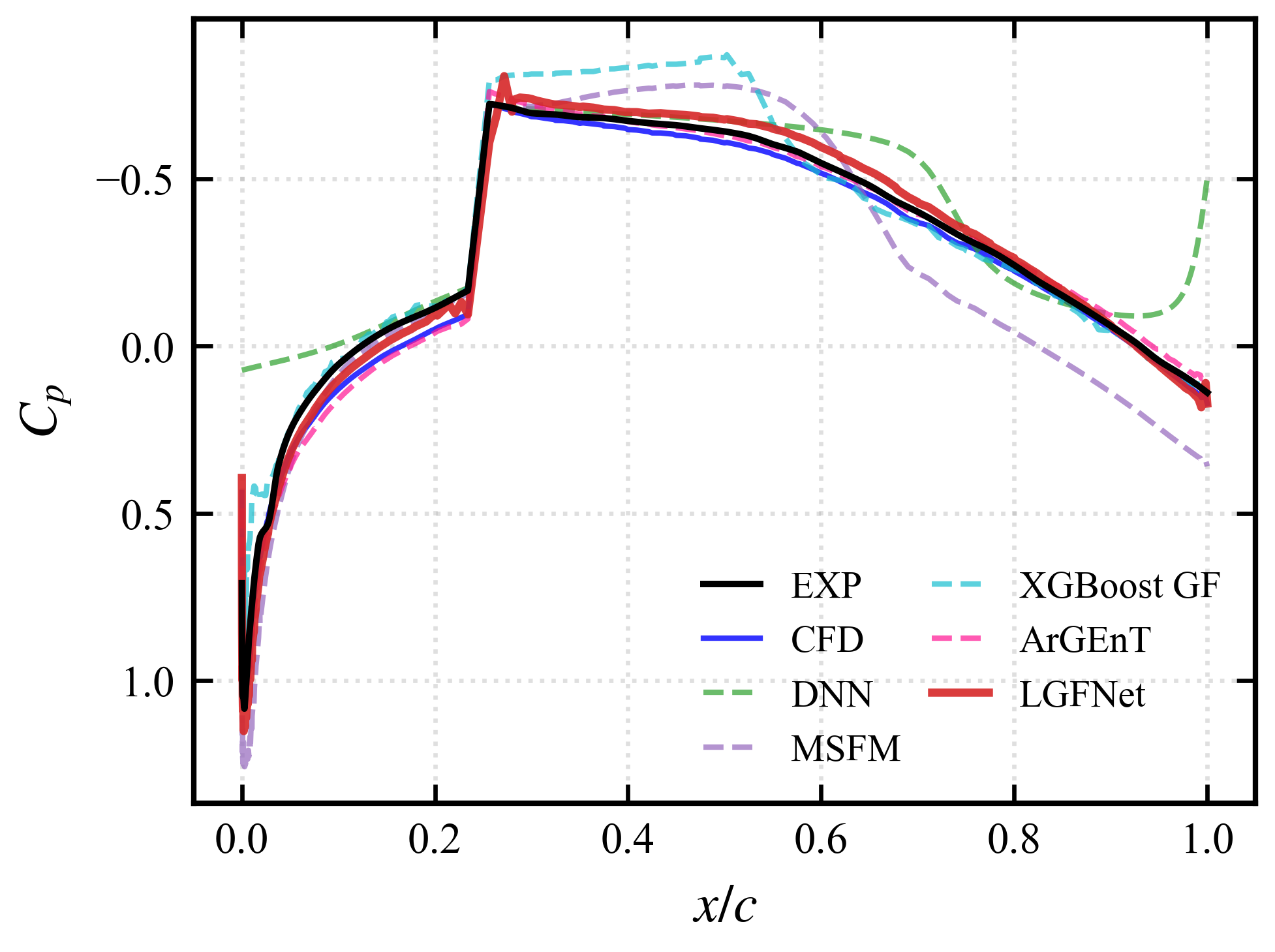} 
        \caption{Case 3: Test Dateset}
        \label{fig:case3_test}
    \end{subfigure}
    
    \vspace{0.5cm} 
    
    \begin{subfigure}[b]{0.32\textwidth}
        \centering
        \includegraphics[width=\linewidth]{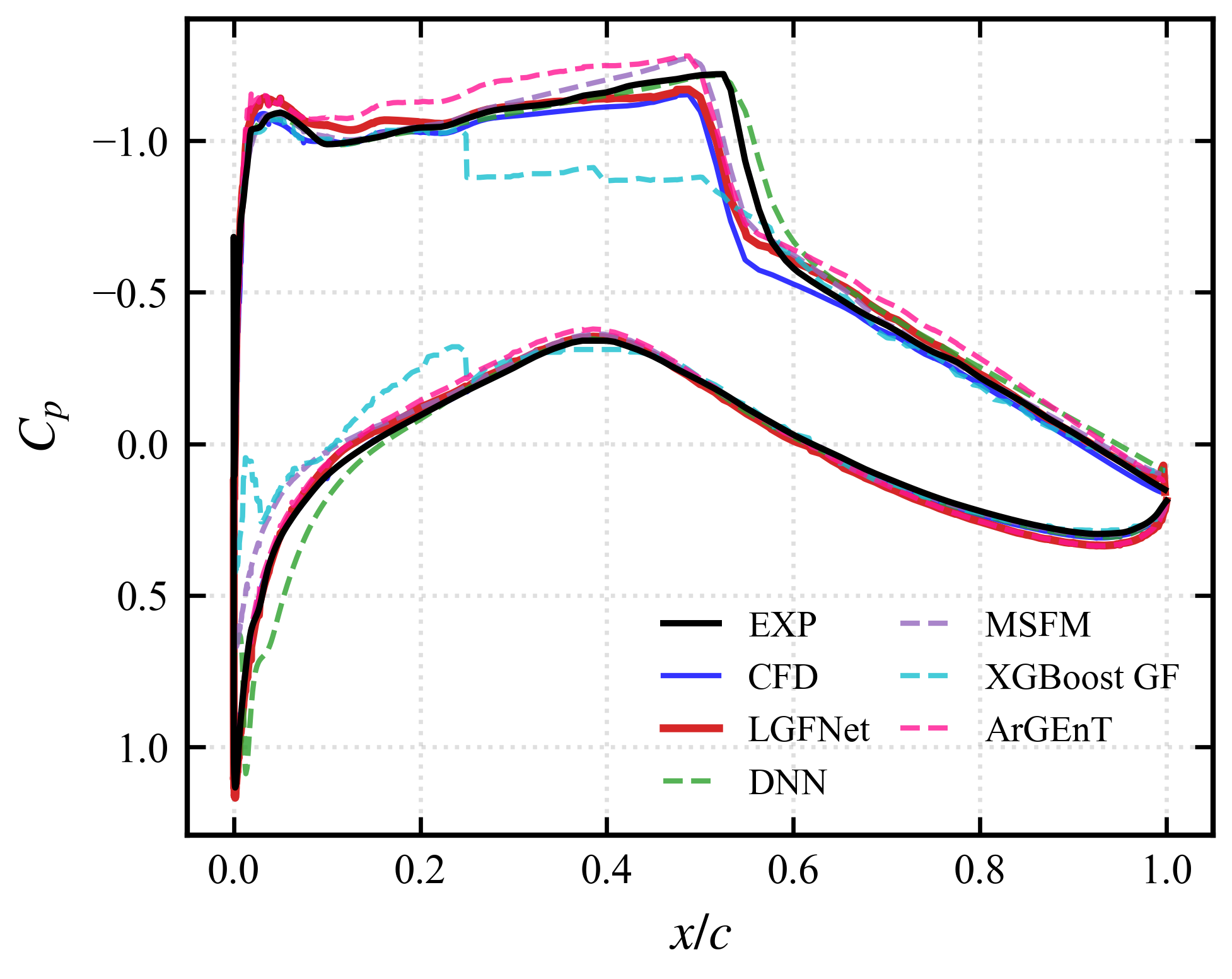} 
        \caption{Case 1: Full Dateset}
		\label{fig:case1_full}
    \end{subfigure}
    \hfill
    \begin{subfigure}[b]{0.32\textwidth}
        \centering
        \includegraphics[width=\linewidth]{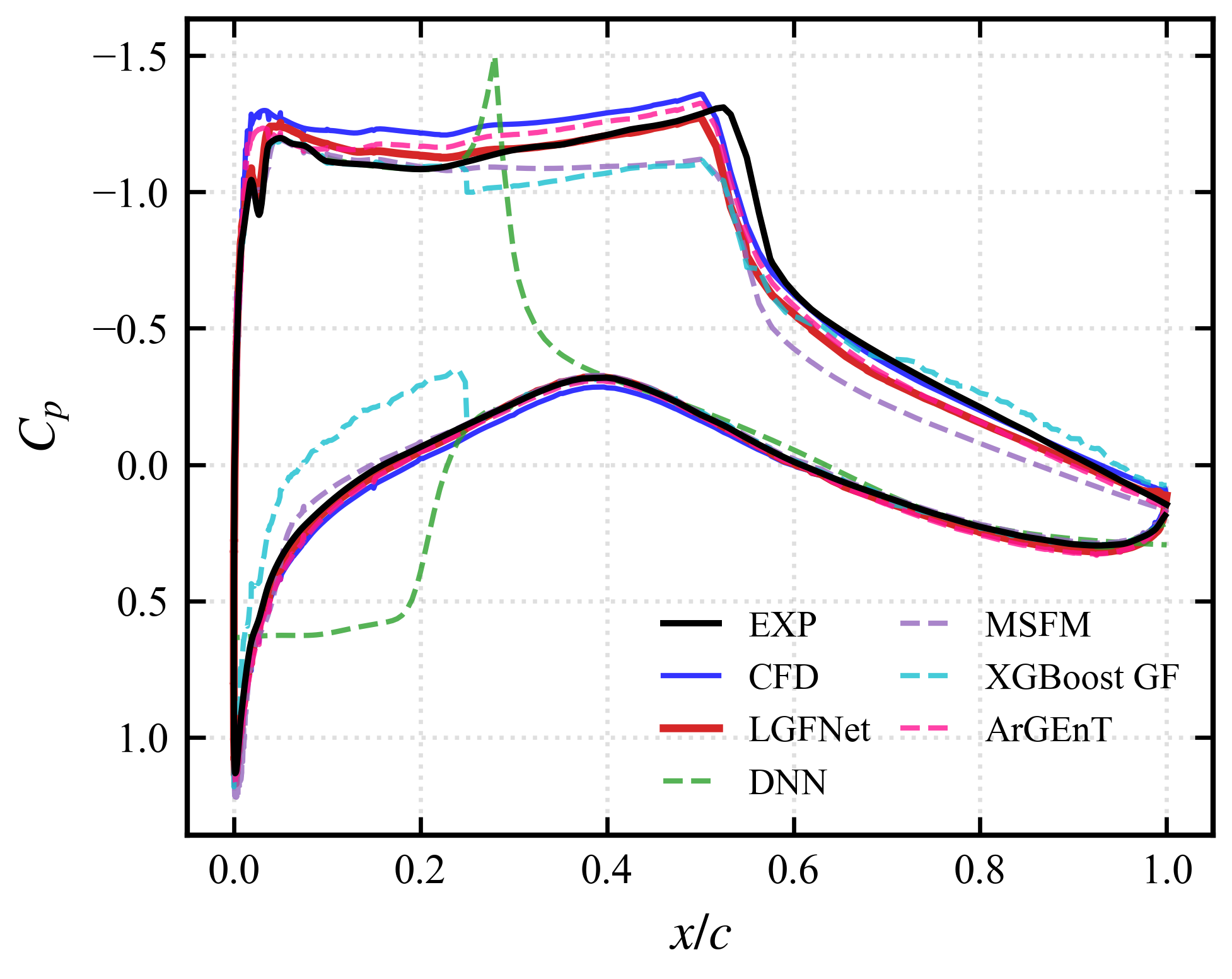} 
        \caption{Case 2: Full Dateset}
		\label{fig:case2_full}
    \end{subfigure}
    \hfill
    \begin{subfigure}[b]{0.32\textwidth}
        \centering
        \includegraphics[width=\linewidth]{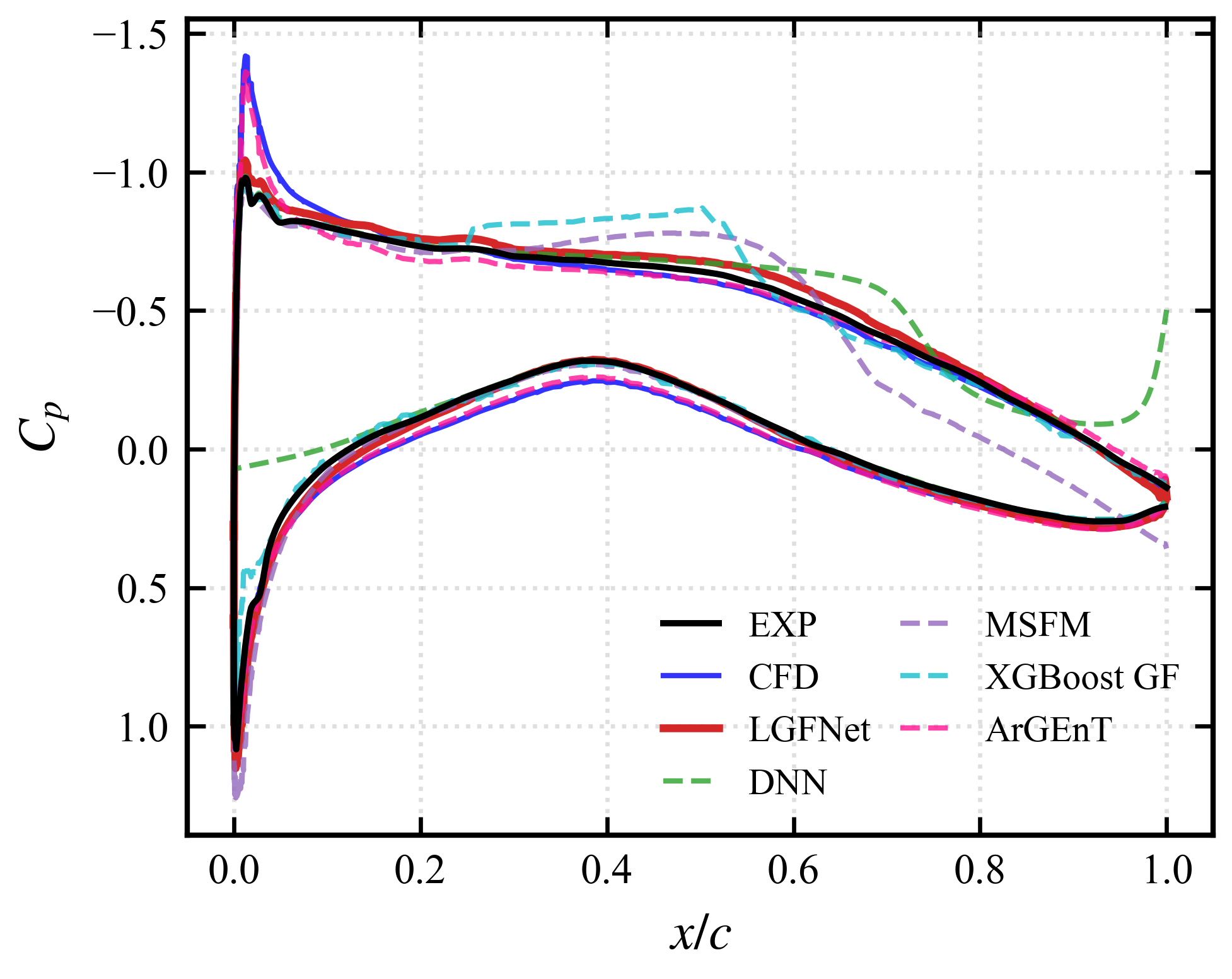} 
		\caption{Case 3: Full Dateset}
        \label{fig:case3_full}
    \end{subfigure}
    
    \caption{Comparison results across three cases. The top row (a-c) shows the pressure coefficient ($C_p$) fusion curves on the test set of the target case, while the bottom row (d-f) displays the fusion curves on the full dataset of the target case.}
    \label{FIG:RAE_Curves}
\end{figure*}

It is worth noting that since HK essentially performs spatial interpolation on the aerodynamic data, its methodology for partitioning and utilizing the training and testing sets differs fundamentally from the data-driven and sliding-window mechanisms of the other baselines in this experiment. Consequently, the visualization of the HK baseline curves is omitted from Fig. \ref{FIG:RAE_Curves}. Its performance, however, is comprehensively summarized through the quantitative metrics in Table \ref{tab:rae_results}. 

\textbf{Quantitative analysis:}
as presented in Table \ref{tab:rae_results}, the LGFNet model consistently achieves the highest prediction accuracy across RMSE, MAE and $R^2$. Taking the subsonic Case 1 as an example, LGFNet achieves an RMSE of 0.0591, significantly reducing the error compared to other baselines. Regarding efficiency, HK and XGBoost exhibit the shortest training times (e.g., HK at 5.24s in Case 2, XGBoost at 5.58s in Case 1) due to their comparatively lightweight structures. Among the deep learning-based methods, ArGEnT shows remarkable training speed (approx. 17-19s) due to its purely attention-based architecture without complex convolutions. However, LGFNet (approx. 103s) is significantly more efficient than DNN (approx. 242s) and comparable to MSFM, achieving an optimal balance between training cost and model performance.

\textbf{Analysis of uncertainty and model reliability:} 
as defined in Section \ref{sec:uncertainty}, the ``Uncertainty'' metric quantifies the model's prediction confidence through the mean width of the 95\% confidence interval. In all cases, the uncertainty of LGFNet is significantly lower than that of the original Wind Tunnel data (e.g., Case 1: 0.1932 vs. 0.7887).This indicates that by integrating the physical trends from CFD, LGFNet effectively filters out noise and reduces the uncertainty inherent in the sparse experimental measurements, resulting in higher confidence predictions.However, a notable anomaly is observed in the transonic Case 2 and Case 3. DNN reports extremely low uncertainty (e.g., Case 3: 0.0547), which is even lower than LGFNet (0.2169), yet its prediction error is excessively high (RMSE 0.3911 vs. 0.0597). This discrepancy suggests that DNN suffers from overconfidence. Lacking the guidance of low-fidelity trends (CFD), DNN fits a simplistic smooth curve through sparse high-fidelity points, failing to detect the complex shock wave structures. Consequently, the GPR model yields an artificially narrow confidence band, severely underestimating the predictive uncertainty in the unmeasured regions.In contrast, LGFNet maintains a reasonable level of uncertainty (much lower than WIND, but higher than the overconfident DNN) while achieving state-of-the-art accuracy. This indicates that LGFNet provides a reliable uncertainty estimate that correctly reflects the complexity of the underlying physics (e.g., shock waves), rather than being blindly confident in an incorrect smooth approximation.

\textbf{Fusion performance across operating cases:}
the fusion curves in Fig. \ref{FIG:RAE_Curves} further visualize the distinct behaviors of each model:
\begin{itemize}
    \item \textbf{Case 1 (Subsonic regime):} The flow field is relatively smooth. While all models capture the general trend, DNN exhibits unphysical oscillations near the leading edge. MSFM and XGBoost produce relatively smooth curves but show slight deviations in the pressure recovery region. LGFNet aligns almost perfectly with the experimental data.
    \item \textbf{Case 2 \& Case 3 (Transonic regime with shock waves):} The two cases involve strong non-linearities, specifically the shock wave discontinuity on the upper surface. 
    \begin{itemize}
        \item DNN completely fails to capture the shock wave, predicting a smooth transition instead of a sharp discontinuity, which explains its high RMSE despite low uncertainty.
        \item MSFM and XGBoost recognize the presence of the shock but suffer from a "smearing effect," where the predicted shock location is offset or the gradient is insufficiently steep compared to the ground truth.
        \item ArGEnT captures the global aerodynamic trends. However, it over-smooths the sharp shock wave discontinuities, exhibiting a smeared gradient that passively follows the low-fidelity CFD input.
        \item LGFNet model accurately reconstructs the shock wave's position and intensity. This proves that Sliding Window mechanism (SW) successfully extracts local high-frequency features (shock discontinuity), while Attention mechanism effectively integrates the global trend from CFD data, correcting the systematic bias while preserving the foundational physical trends of the CFD baseline.
    \end{itemize}
\end{itemize}

LGFNet demonstrates superior robustness and fidelity, effectively correcting low-fidelity data to match high-fidelity experiments even in the presence of complex flow discontinuities.

\begin{table}[ht]
\centering
\caption{Quantitative comparison of four fusion models on RAE2822 dataset.}\label{tab:rae_results}
\resizebox{\linewidth}{!}{%
\begin{tabular}{llccccc}
\toprule
Case & Model & RMSE $\downarrow$ & MAE $\downarrow$ & $R^2$ $\uparrow$ & Time(s) & Uncertainty $\downarrow$ \\
\midrule
\multirow{5}{*}{Case 1} 
& WIND (Exp.) & - & - & - & - & 0.7887 \\
& HK & 0.1503 & 0.09769 & 0.9422 & 6.08 & 0.2852 \\
& DNN & 0.1704 & 0.1165 & 0.9290 & 242.60 & 0.2341 \\
& MSFM & 0.1682 & 0.1092 & 0.9308 & 118.99 & 0.2648 \\
& XGBoost & 0.3256 & 0.2216 & 0.7410 & \textbf{5.58} & 0.2956 \\
& ArGEnT & 0.0681 & 0.0552 & 0.9887 & 17.02 & 0.1695 \\
& \textbf{LGFNet (Ours)} & \textbf{0.0591} & \textbf{0.0394} & \textbf{0.9915} & 103.33 & \textbf{0.1541} \\
\midrule
\multirow{5}{*}{Case 2} 
& WIND (Exp.) & - & - & - & - & 0.6314 \\
& HK & 0.1887 & 0.1120 & 0.9182 & \textbf{5.24} & 0.26595 \\
& DNN & 0.4719 & 0.4049 & 0.4899 & 246.13 & 0.1503 \\
& MSFM & 0.1104 & 0.0892 & 0.9721 & 110.10 & 0.2593 \\
& XGBoost & 0.1579 & 0.1356 & 0.9429 & 6.90 & 0.1860 \\
& ArGEnT & 0.0690 & 0.0593 & 0.9891 & 17.16 & 0.1687 \\
& \textbf{LGFNet (Ours)} & \textbf{0.0607} & \textbf{0.0411} & \textbf{0.9916} & 103.25 & \textbf{0.1240} \\
\midrule
\multirow{5}{*}{Case 3} 
& WIND (Exp.) & - & - & - & - & 0.6618 \\
& HK & 0.2043 & 0.1151 & 0.8454 & \textbf{6.47} & 0.27504 \\
& DNN & 0.3911 & 0.2481 & 0.3982 & 242.24 & \textbf{0.0547} \\
& MSFM & 0.1618 & 0.1355 & 0.8970 & 136.95 & 0.2553 \\
& XGBoost & 0.1118 & 0.0706 & 0.9508 & 6.55 & 0.2020 \\
& ArGEnT & 0.0707 & 0.0538 & 0.9804 & 19.10 & 0.2006 \\
& \textbf{LGFNet (Ours)} & \textbf{0.0597} & \textbf{0.0457} & \textbf{0.9860} & 108.49 & 0.1945 \\
\bottomrule
\end{tabular}
}
\end{table}

\subsubsection{Results on CARDC aircraft dataset}
\begin{figure*}[htbp]
    \centering
    \begin{subfigure}[b]{0.48\textwidth}
        \centering
        \includegraphics[width=\linewidth]{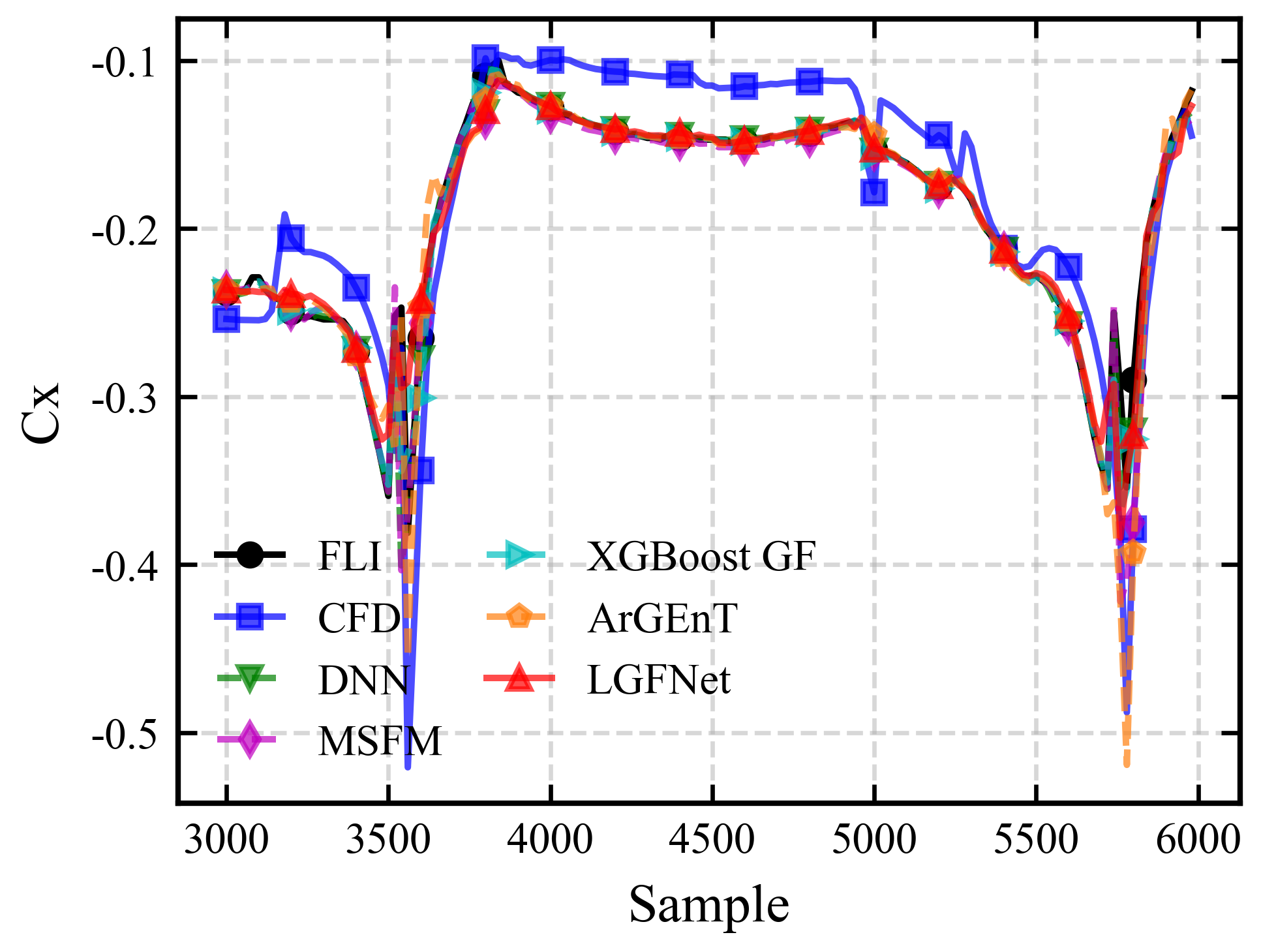} 
        \caption{Fusion on sample segments: Cx}
        \label{fig:Cx}
    \end{subfigure}
    \hfill 
    \begin{subfigure}[b]{0.48\textwidth}
        \centering
        \includegraphics[width=\linewidth]{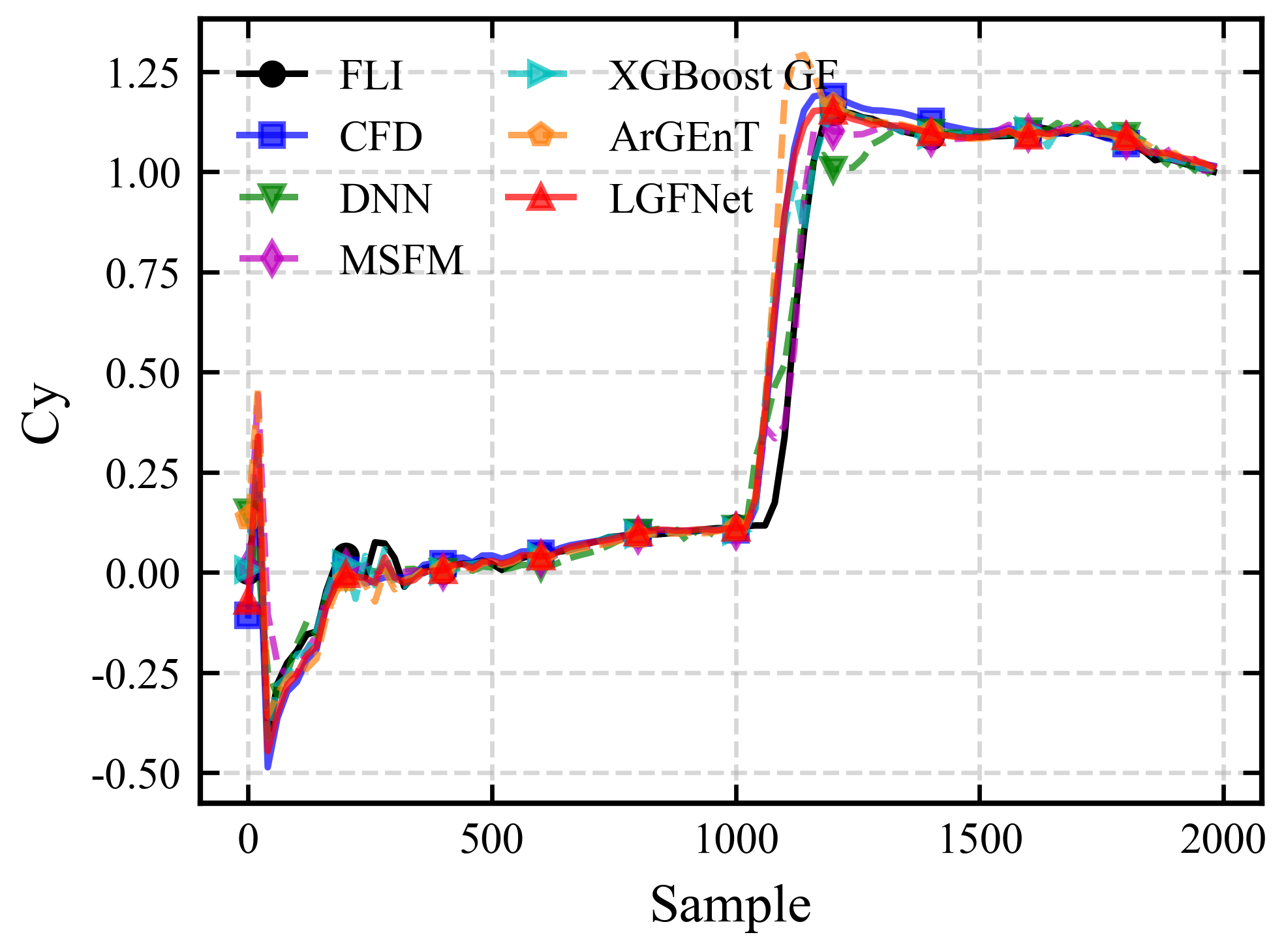} 
        \caption{Fusion on sample segments: Cy}
        \label{fig:Cy}
    \end{subfigure}
    
    \vspace{0.5cm} 
    
    \begin{subfigure}[b]{0.48\textwidth}
        \centering
        \includegraphics[width=\linewidth]{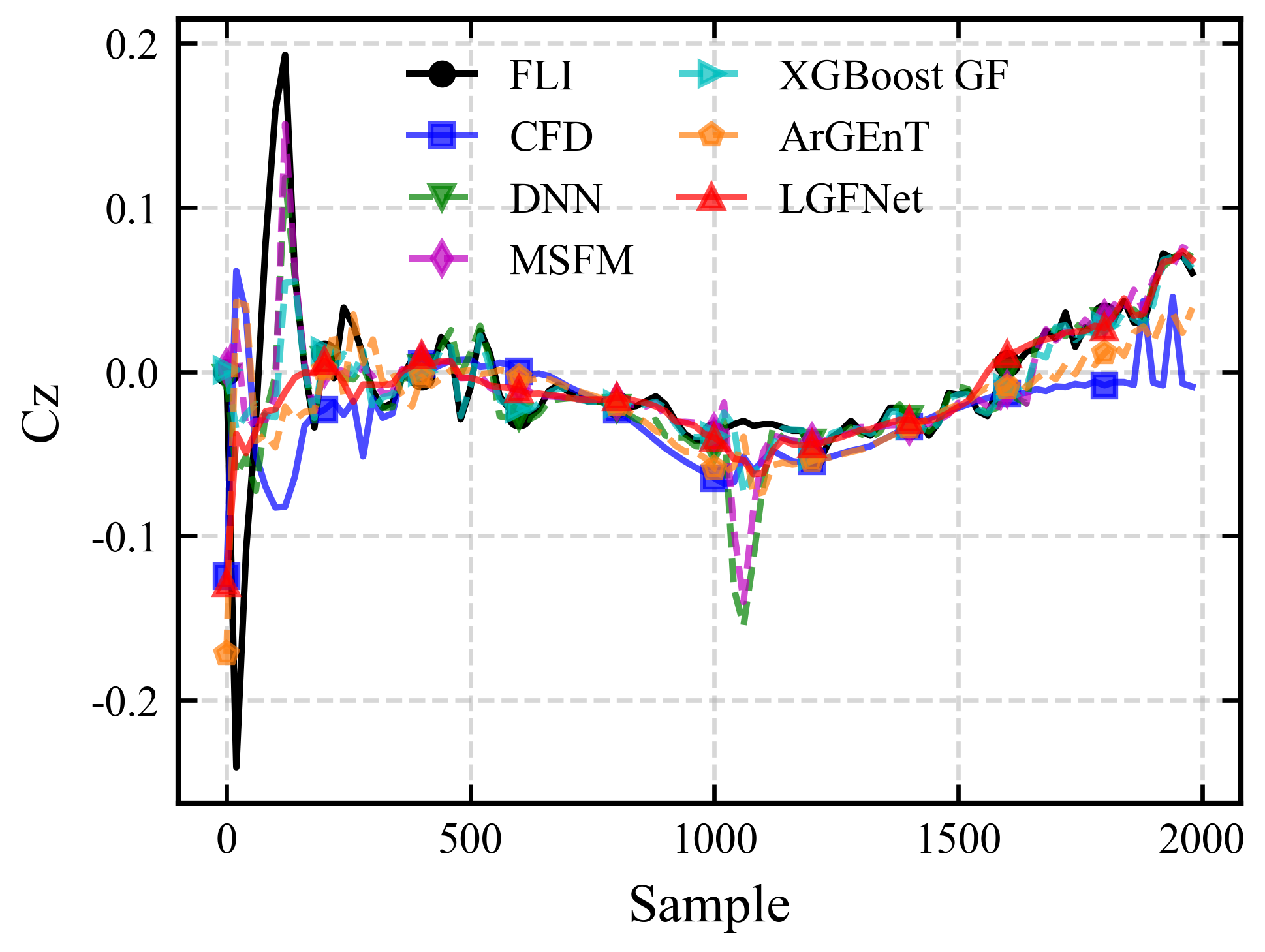} 
        \caption{Fusion on sample segments: Cz}
		\label{fig:Cz1}
    \end{subfigure}
    \hfill
    \begin{subfigure}[b]{0.48\textwidth}
            \centering
            \includegraphics[width=\linewidth]{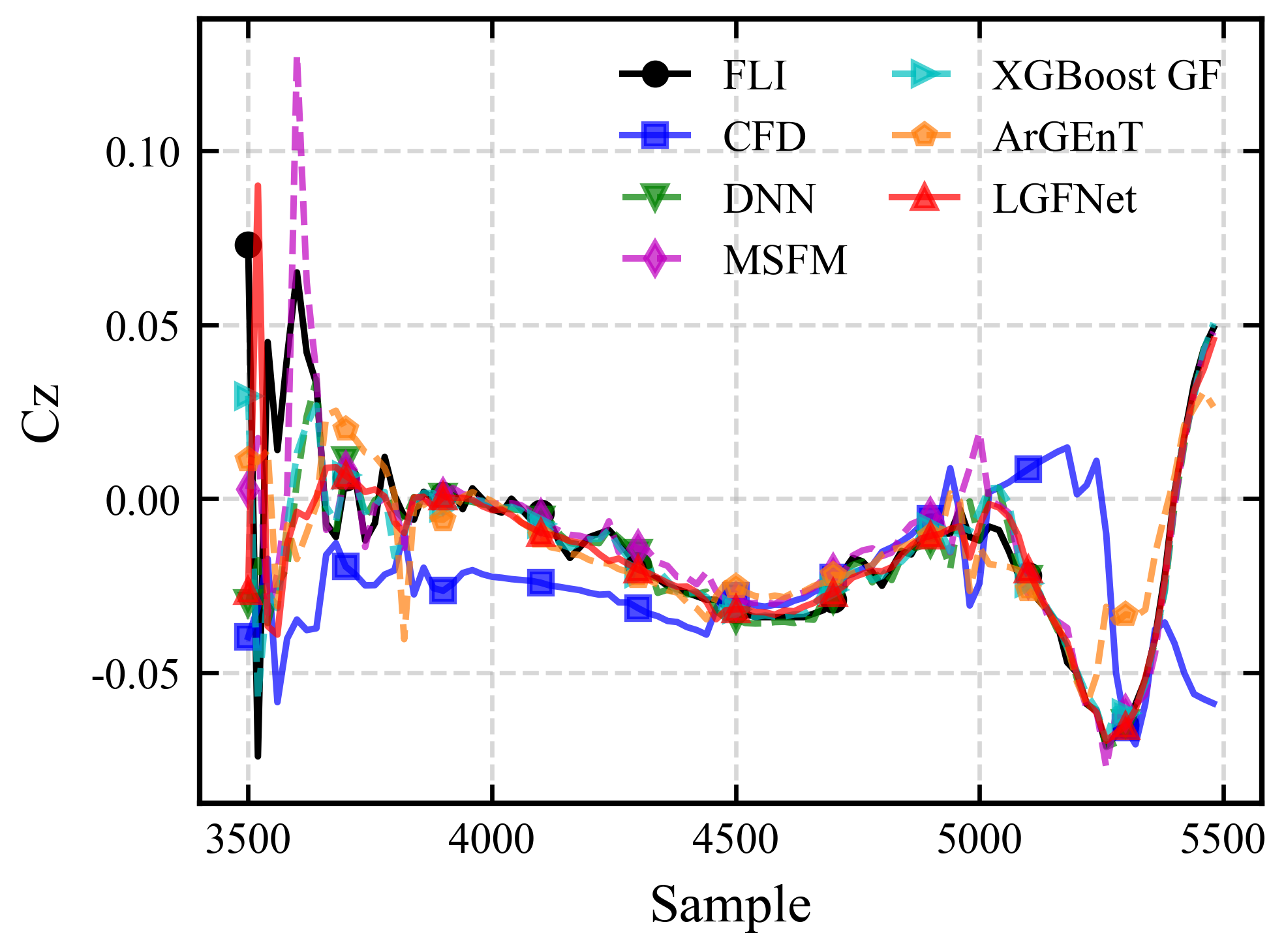} 
            \caption{Fusion on sample segments: Cz}
            \label{fig:Cz2}
    \end{subfigure}
    
    \caption{Comparison results of different aerodynamic coefficients.}
    \label{FIG:CADRC_Curves}
\end{figure*}

To evaluate the engineering applicability of LGFNet, comparative experiments were conducted on the CARDC aircraft dataset, focusing on the prediction of concentrated force coefficients ($C_{x}, C_{y}, C_{z}$) across a high-dimensional flight envelope. It is worth noting that the Hierarchical Kriging (HK), which was evaluated in Scenario 1, is excluded from this baseline comparison. This is because traditional Gaussian Process-based methods like HK suffer from $\mathcal{O}(N^3)$ computational complexity, making them computationally prohibitive and prone to convergence issues when applied to the large-scale and high-dimensional state space of the CARDC dataset. The quantitative performance of the remaining models is summarized in Table \ref{tab:cardc_results}. Furthermore, Fig \ref{FIG:CADRC_Curves} illustrates the fusion curves of various algorithm models (including DNN, MSFM, XGBoost GF, ArGEnT and LGFNet) on the same sample segments for each aerodynamic coefficient, providing a direct visual comparison of their modeling characteristics.

\textbf{Quantitative performance and uncertainty analysis:} evaluated using the GPR-based metric described in Section \ref{sec:uncertainty}, the experimental data reveal that the raw FLI measurements possess the highest uncertainty across all three coefficients (e.g., $C_z$: 1.0786). Notably, all fusion models significantly reduce this uncertainty metric, effectively enhancing the overall reliability of the aerodynamic data. Among the baselines, XGBoost GF exhibits the shortest training time (approx. 4s) due to its tree-based structure. Regarding $C_x$ and $C_y$, LGFNet maintains high precision with $R^2$ values exceeding 0.984. In contrast, while ArGEnT performs reasonably well on the relatively smooth $C_x$ coefficient ($R^2$ 0.9850), its accuracy drops significantly on $C_y$ (yielding the highest RMSE of 0.1735 and lowest $R^2$ of 0.9506). This performance degradation indicates that a pure cross-attention mechanism, devoid of local spatial inductive biases like sliding windows, struggles to adequately capture the complex high-frequency cross-coupling effects inherent in normal force prediction.However, for the $C_y$ coefficient, LGFNet's RMSE (0.0964) is higher than that of MSFM (0.0353). This gap likely stems from the lightweight channel configuration [8, 16, 32, 64, 128] adopted for Scenario 2 to prevent overfitting, which may limit the model's capacity to represent the complex cross-coupling effects inherent in normal force $(C_y)$ within this specific data regime. Future hyperparameter tuning of the encoder width could further bridge this discrepancy. However, the advantage of LGFNet is most prominent in the $C_z$ coefficient, where it achieves the lowest RMSE (0.0169), the highest $R^2$ (0.9344), and the lowest uncertainty (0.1998) among all models.

\textbf{Analysis of fusion curves:} the visual results on the same sample segments reveal the distinct behaviors of the models:
\begin{itemize}
    \item \textbf{Axial force ($C_x$) and normal force ($C_y$):} in these scenarios, LGFNet, MSFM, DNN, XGBoost and ArGEnT show highly consistent fusion performance, successfully bridging the systematic bias between the CFD trends and high-fidelity FLI data points.  LGFNet tracks the complex fluctuations and sharp jumps of the $C_y$ coefficient (e.g., in the vicinity of samples 1000–1200 in Fig \ref{fig:Cy}) with high fidelity, showing no significant deviation from the other advanced deep learning baselines.
    \item \textbf{Lateral force ($C_z$):} the $C_z$ coefficient presents the most challenging fusion task due to high-frequency noise and sparse distribution. While DNN and MSFM models exhibit counter-trend inverse oscillations (e.g., in the vicinity of samples 1000–1200 in Fig \ref{fig:Cz1}), and ArGEnT struggles slightly to fully suppress high-frequency local noise without a sliding window, LGFNet maintains global coherence. It provides a much smoother trajectory that accurately captures the essential physical transitions of the FLI data while strictly adhering to the underlying physical trends reflected in the CFD simulations.
\end{itemize}

LGFNet demonstrates robust generalization across the CARDC dataset. Its ability to integrate local features and global dependencies provides a decisive advantage in processing the highly noisy $C_z$ measurements, resulting in a more reliable aerodynamic database that strictly adheres to the baseline physical trends.

\begin{table}[ht]
\centering
\caption{Quantitative comparison of four fusion models on CARDC aircraft dataset.}\label{tab:cardc_results}
\resizebox{\linewidth}{!}{%
\begin{tabular}{llccccc}
\toprule
Case & Model & RMSE $\downarrow$ & MAE $\downarrow$ & $R^2$ $\uparrow$ & Time(s) & Uncertainty $\downarrow$ \\
\midrule
\multirow{5}{*}{$C_x$} 
& FLI (Exp.) & - & - & - & - & 0.3394 \\
& DNN & \textbf{0.0037} & \textbf{0.0023} & \textbf{0.9923} & 188.84 & 0.1948 \\
& MSFM & 0.0040 & 0.0025 & 0.9909 & 668.26 & 0.2078 \\
& XGBoost & 0.0043 & 0.0028 & 0.9892 & \textbf{2.43} & \textbf{0.1943} \\
& ArGEnT & 0.0042 & 0.0032 & 0.9850 & 58.22 & 0.2453 \\
& \textbf{LGFNet (Ours)} & 0.0042 & 0.0027 & 0.9848 & 243.88 & 0.1953 \\
\midrule
\multirow{5}{*}{$C_y$} 
& FLI (Exp.) & - & - & - & - & 2.1005 \\
& DNN & 0.0660 & 0.0314 & 0.9925 & 215.43 & 1.7268 \\
& \textbf{MSFM} & \textbf{0.0353} & \textbf{0.0175} & \textbf{0.9978} & 403.37 & \textbf{1.3084} \\
& XGBoost & 0.0947 & 0.0403 & 0.9846 & \textbf{4.12} & 1.7720 \\
& ArGEnT & 0.1735 & 0.0509 & 0.9506 & 66.97 &  1.7240\\
& LGFNet & 0.0964 & 0.0355 & 0.9847 & 226.43 & 1.7712 \\
\midrule
\multirow{5}{*}{$C_z$} 
& FLI (Exp.) & - & - & - & - & 1.0786 \\
& DNN & 0.0234 & 0.0120 & 0.8662 & 203.32 & 0.2075 \\
& MSFM & 0.0233 & 0.0131 & 0.8670 & 638.73 & 0.2087 \\
& XGBoost & 0.0172 & 0.0116 & 0.9279 & \textbf{4.14} & 0.2047 \\
& ArGEnT & 0.0198 & 0.0139 & 0.9104 & 46.96 & 0.3414 \\
& \textbf{LGFNet (Ours)} & \textbf{0.0169} & \textbf{0.0108} & \textbf{0.9344} & 241.32 & \textbf{0.1998} \\
\bottomrule
\end{tabular}
}
\end{table}

\subsection{Ablation experiments}
To rigorously evaluate the individual contributions of the core components within the proposed LGFNet framework, we conduct a series of ablation studies across both datasets. Three distinct model variants are examined: (1) \textbf{LGFNet (full model)}, (2) \textbf{LGFNet w/o SW \& Att} (both modules removed), (3) \textbf{LGFNet w/o SW} (sliding window removed), and (4) \textbf{LGFNet w/o Att} (self-attention mechanism removed). The quantitative results are summarized in terms of RMSE, MAE, and $R^2$. The quantitative results are summarized in Tables \ref{tab:ablation_rae} and \ref{tab:ablation_cardc}.

\subsubsection{Analysis on RAE2822 airfoil dataset}
For the pressure distribution task on the RAE2822 dataset, a consistent performance trend is observed across all operating cases. The full LGFNet model exhibits a significant superiority over all ablation variants.
\begin{itemize}
    \item The SW module provides a critical local inductive bias. By establishing a localized receptive field over the spatial sequences, SW enhances the model's ability to capture nonlinear jumps in the $C_p$ curves. Specifically, in Case 2, the removal of SW leads to an increase in RMSE from 0.0607 to 0.0717. This demonstrates that SW is indispensable for maintaining local physical gradients and preventing numerical oscillations near shock wave positions.
    \item The Att module is responsible for correlating global physical features. Across all cases, the integration of Att improves the $R^2$ metric. In Case 1, the full model achieves the highest correlation ($R^2=0.9915$), proving that the self-attention mechanism can leverage global context to calibrate systematic biases in CFD data, ensuring the fused results adhere to holistic aerodynamic constraints.
\end{itemize}

\begin{figure*}[htbp]
    \centering
    \begin{subfigure}[b]{0.32\textwidth}
        \centering
        \includegraphics[width=\linewidth]{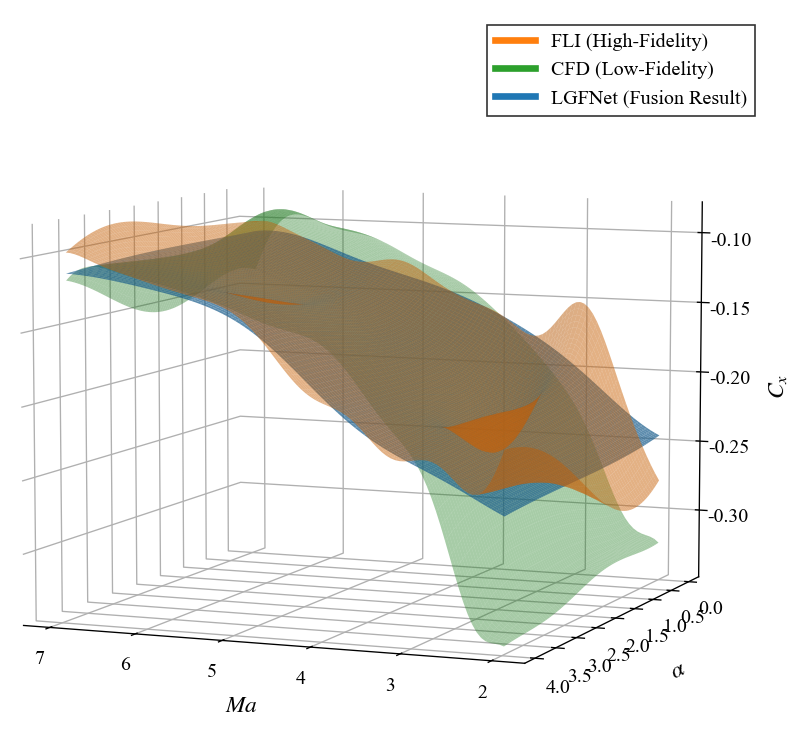} 
        \caption{Cx}
        \label{fig:Cx_3d}
    \end{subfigure}
    \hfill 
    \begin{subfigure}[b]{0.32\textwidth}
        \centering
        \includegraphics[width=\linewidth]{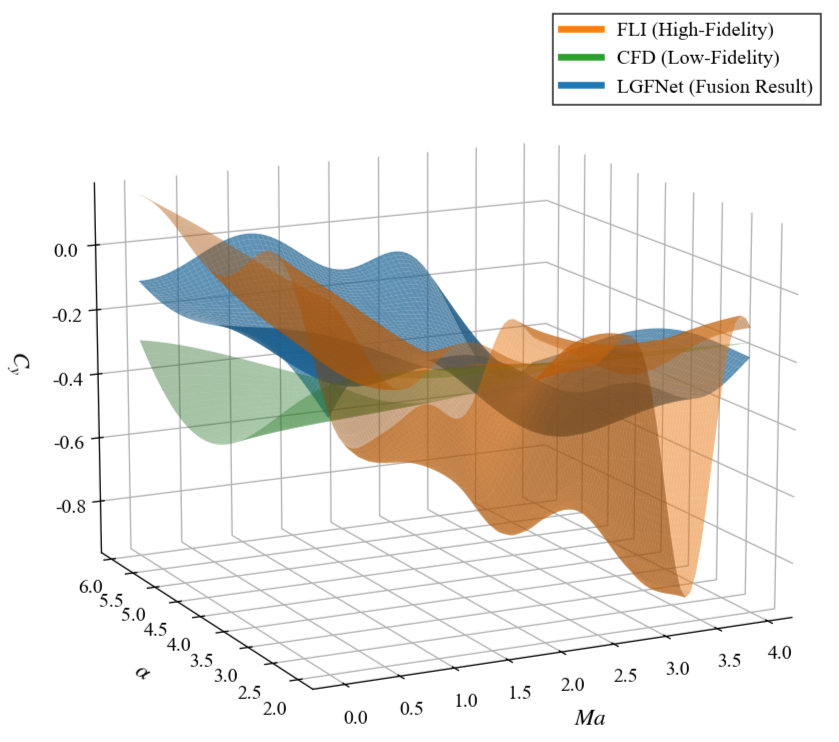} 
        \caption{Cy}
        \label{fig:Cy_3d}
    \end{subfigure}
    \hfill
    \begin{subfigure}[b]{0.32\textwidth}
        \centering
        \includegraphics[width=\linewidth]{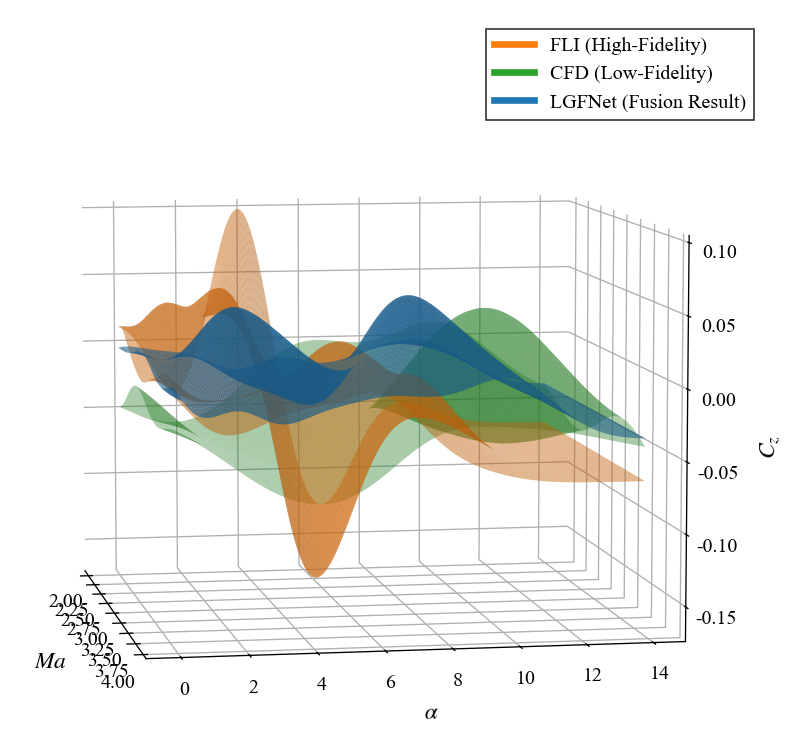} 
        \caption{Cz}
        \label{fig:Cz_3d}
    \end{subfigure}
    
    \caption{Surface comparison of FLI,CFD and LGFNet with respect to aerodynamic coefficients.}
    \label{FIG:3d_Curves}
\end{figure*}

\begin{table}[ht]
\centering
\caption{Ablation study results on RAE2822 airfoil dataset.}
\label{tab:ablation_rae}
\begin{tabular*}{\tblwidth}{@{} LCCCC@{} }
\toprule
Model Variant & Case & RMSE & MAE & $R^2$ \\
\midrule
LGFNet (w/o SW \& Att) & Case 1 & 0.0892 & 0.0758 & 0.9805 \\
LGFNet (w/o SW)        & Case 1 & 0.0821 & 0.0559 & 0.9835 \\
LGFNet (w/o Att)       & Case 1 & 0.0791 & 0.0551 & 0.9847 \\
\textbf{LGFNet (Full)} & Case 1 & \textbf{0.0591} & \textbf{0.0394} & \textbf{0.9915} \\
\midrule
LGFNet (w/o SW \& Att) & Case 2 & 0.0846 & 0.0763 & 0.9835 \\
LGFNet (w/o SW)        & Case 2 & 0.0717 & 0.0512 & 0.9882 \\
LGFNet (w/o Att)       & Case 2 & 0.0609 & 0.0456 & 0.9915 \\
\textbf{LGFNet (Full)} & Case 2 & \textbf{0.0607} & \textbf{0.0411} & \textbf{0.9916} \\
\midrule
LGFNet (w/o SW \& Att) & Case 3 & 0.0987 & 0.0822 & 0.9754 \\
LGFNet (w/o SW)        & Case 3 & 0.0854 & 0.0614 & 0.9812 \\
LGFNet (w/o Att)       & Case 3 & 0.0802 & 0.0588 & 0.9841 \\
\textbf{LGFNet (Full)} & Case 3 & \textbf{0.0597} & \textbf{0.0402} & \textbf{0.9908} \\
\bottomrule
\end{tabular*}
\end{table}   

\subsubsection{Analysis on CARDC aircraft dataset}
In predicting high-dimensional aerodynamic coefficients for the CARDC dataset, the task shifts toward extracting stable aerodynamic characteristics from sparse and noisy flight test samples. Generally, the SW module acts as a "manifold smoother" by introducing overlapping sequences that augment the input state representation. Meanwhile, the Att module drastically enhances the nonlinear mapping capability by learning the complex inter-dependencies between flight state variables (e.g., $Ma, \alpha, \phi$).

Among the three coefficients, $C_z$ (lateral force) presents the greatest fusion challenge due to its small magnitude and high sensitivity to stochastic disturbances such as crosswinds. The results show that the full LGFNet significantly reduces the RMSE (0.0169) compared to variants lacking Att or SW. This proves that the Att module functions as a "physical filter" in $C_z$ prediction, suppressing high-frequency noise by prioritizing features aligned with the global aerodynamic state. Simultaneously, SW improves the perception of local features under complex sideslip conditions, ensuring the robustness of the fusion process for the most challenging coefficient.

\begin{table}[ht]
\centering
\caption{Ablation study results on CARDC aircraft dataset.}
\label{tab:ablation_cardc}
\begin{tabular*}{\tblwidth}{@{} LCCCC@{} }
\toprule
Model Variant & Coeff. & RMSE & MAE & $R^2$ \\
\midrule
LGFNet (w/o SW \& Att) & $C_x$ & 0.0075 & 0.0053 & 0.9752 \\
LGFNet (w/o SW)        & $C_x$ & 0.0058 & 0.0042 & 0.9789 \\
LGFNet (w/o Att)       & $C_x$ & 0.0056 & 0.0039 & 0.9812 \\
\textbf{LGFNet (Full)} & $C_x$ & \textbf{0.0042} & \textbf{0.0031} & \textbf{0.9848} \\
\midrule
LGFNet (w/o SW \& Att) & $C_y$ & 0.0094 & 0.0076 & 0.9654 \\
LGFNet (w/o SW)        & $C_y$ & 0.0082 & 0.0061 & 0.9712 \\
LGFNet (w/o Att)       & $C_y$ & 0.0079 & 0.0058 & 0.9745 \\
\textbf{LGFNet (Full)} & $C_y$ & \textbf{0.0063} & \textbf{0.0045} & \textbf{0.9811} \\
\midrule
LGFNet (w/o SW \& Att) & $C_z$ & 0.0214 & 0.0182 & 0.9215 \\
LGFNet (w/o SW)        & $C_z$ & 0.0195 & 0.0165 & 0.9402 \\
LGFNet (w/o Att)       & $C_z$ & 0.0188 & 0.0159 & 0.9455 \\
\textbf{LGFNet (Full)} & $C_z$ & \textbf{0.0169} & \textbf{0.0135} & \textbf{0.9588} \\
\bottomrule
\end{tabular*}
\end{table}

\subsection{3D visualization experiment}
To intuitively evaluate the model's generalization capability across the global flight envelope and verify the preservation of underlying physical trends in the fused results, we conducted a 3D visualization experiment on the CARDC aircraft dataset (Scenario 2).

Since the original flight test data (FLI) and CFD data comprise discrete points distributed in a high-dimensional space, they cannot directly form continuous surfaces for intuitive comparison. To visualize the aerodynamic laws effectively, we employed Gaussian Process Regression (GPR) with a squared-exponential kernel to construct continuous response surfaces. Specifically, we selected Mach number (Ma) and Angle of Attack ($\alpha$) as the independent variables (x and y axes), while fixing other state variables. We then utilized the trained GPR models to predict the aerodynamic responses on a dense, uniform meshgrid, thereby generating the continuous surfaces for Low-Fidelity (CFD), High-Fidelity (FLI), and the LGFNet Fusion results,as shown in Fig. \ref{FIG:3d_Curves}.

Fig. \ref{FIG:3d_Curves} presents the surface comparisons for the axial force ($C_x$), normal force ($C_y$), and lateral force ($C_z$) coefficients. Green Surface (CFD) represents the low-fidelity simulation data. Although it covers the entire envelope, there is a clear "systematic bias" (offset) compared to the orange high-fidelity surface, indicating lower numerical accuracy. Orange Surface (FLI) represents the high-fidelity flight test data. While accurate, the surface exhibits local roughness and discontinuities due to measurement noise and sparse sampling (especially visible in Fig. \ref{fig:Cz_3d}). Blue Surface (LGFNet) represents the fusion result of our proposed model.

As observed in the visualization, the LGFNet fusion surface (Blue) tightly adheres to the high-fidelity FLI surface (Orange) in the vertical value dimension, significantly correcting the systematic errors present in the CFD data. More importantly, regarding the geometric topology, the LGFNet surface inherits the smoothness and trend variations of the CFD surface (Green). For instance, in the nonlinear variation regions of $C_y$ (Fig. \ref{fig:Cy_3d}), LGFNet model effectively suppresses the high-frequency noise found in the FLI data while preserving the physical gradients captured by CFD.

\section{Conclusion}

In this paper, we propose LGFNet to address the challenges of multi-source aerodynamic data fusion. The model proposes a synergistic local-global architecture that captures fine-grained local features and long-range flow information. 
Experiments demonstrate that LGFNet achieves state-of-the-art performance, notably reducing RMSE by approximately 65\% in RAE2822 airfoil scenarios while precisely reconstructing sharp shock wave discontinuities. In high-dimensional CARDC aircraft coefficient fusion, the model demonstrates superior robustness, particularly for the lateral force ($C_z$) by achieving the lowest RMSE (0.0169).

Looking ahead, future research will focus on two key directions. First, we aim to investigate advanced encoding mechanisms within the RRL to better reflect the heterogeneous physical contributions of different aerodynamic feature dimensions. Second, considering the $O(N^2)$ computational complexity inherent in standard softmax-based self-attention mechanisms, we plan to integrate efficient attention variants into the RRL,so as to significantly reduce the memory footprint and inference latency. 

\section*{Acknowledgments}
The authors would like to express their sincere gratitude to the China Aerodynamics Research and Development Center (CARDC) for their great assistance and support in data acquisition.

\section*{Data availability}
Data will be made available on request.

\bibliographystyle{model1-num-names}

\bibliography{cas-refs}

@article{Kumar2025,
  author = {Kumar, J. K. A. and Prabhu, K. S. and Veerendra, A. S. and others},
  title = {A comprehensive review of aerodynamic performance evaluation for aircraft wings and engines: insights into experimental, numerical, and theoretical approaches},
  journal = {Aerospace Systems},
  year = {2025},
  doi = {10.1007/s42401-025-00366-w},
  url = {https://doi.org/10.1007/s42401-025-00366-w}
}

@INPROCEEDINGS{Chauhan2017,
  author={Chauhan, R. K. and Singh, S.},
  booktitle={2017 International Conference on Infocom Technologies and Unmanned Systems (Trends and Future Directions) (ICTUS)}, 
  title={Review of aerodynamic parameter estimation techniques}, 
  year={2017},
  volume={},
  number={},
  pages={864-869},
  doi={10.1109/ICTUS.2017.8286127}}

@article{Ding2025,
author = {Ding, Di and Wang, Qing and Liu, Jin and Luo, Wei and Wang, An},
title = {A Joint Online Estimation Method for Aircraft Aerodynamic Parameters and Thrust Deviation},
journal = {International Journal of Aerospace Engineering},
volume = {2025},
number = {1},
pages = {1890214},
doi = {https://doi.org/10.1155/ijae/1890214},
url = {https://onlinelibrary.wiley.com/doi/abs/10.1155/ijae/1890214},
eprint = {https://onlinelibrary.wiley.com/doi/pdf/10.1155/ijae/1890214},
year = {2025}
}

@ARTICLE{Tang2023,
  author  = {Tang, Z. G. and Yuan, X. X. and Qian, W. Q. and others},
  title   = {Research progress on the fusion of data obtained by high-speed wind tunnels, {CFD} and model flight},
  journal = {Acta Aerodynamica Sinica},
  volume  = {41},
  number  = {8},
  year    = {2023},
  pages   = {44--58},
  doi     = {10.7638/kqdlxxb-2023.0096}
}

@INPROCEEDINGS{Poloczek2017,
  author  = {Poloczek, M. and Wang, J. and Frazier, P. I.},
  title   = {Multi information source optimization},
  booktitle = {Proceedings of the 31st International Conference on Neural Information Processing Systems},
  publisher = {Curran Associates Inc.},
  address = {Long Beach, California, USA},
  year    = {2017},
  pages   = {4291--4301}
}

@ARTICLE{Hu2025,
  author  = {Hu, D. F. and Xiang, Y. and Zhang, J. and others},
  title   = {Multi source data fusion method based on radial basis function generative adversarial network},
  journal = {Acta Aeronautica et Astronautica Sinica},
  volume  = {46},
  number  = {10},
  year    = {2025},
  pages   = {631478},
  doi     = {10.7527/S1000-6893.2025.31478}
}

@ARTICLE{Qiu2024,
  author  = {Qiu, J. X. and Si, H. Q. and Gao, X. R. and others},
  title   = {Aerodynamic data fusion method under epistemic uncertainty in flight tests},
  journal = {Acta Aerodynamica Sinica},
  volume  = {42},
  number  = {10},
  year    = {2024},
  pages   = {69--83},
  doi     = {10.7638/kqdlxxb-2024.0109}
}

@ARTICLE{Wang2025,
  author  = {Wang, P. F. and Zeng, L. F. and Shao, X. M. and others},
  title   = {Research on {Multi-source} {Data} {Fusion} {Modeling} {Method} for {Aerodynamic} {Load} of {Aircraft} {Wing} {Based} on {Pre-training} and {Fine-tuning}},
  journal = {Acta Aeronautica et Astronautica Sinica},
  volume  = {46},
  year    = {2025},
  doi     = {10.7527/S1000-6893.2025.32297}
}

@ARTICLE{He2014,
  author  = {He, K. F. and Qian, W. Q. and Wang, Q. and others},
  title   = {Application of data fusion technique in aerodynamics studies},
  journal = {ACTA Aerodynamica Sinica},
  volume  = {32},
  number  = {6},
  year    = {2014},
  pages   = {777--782}
}

@article{Brunton_2020,
   title={Machine Learning for Fluid Mechanics},
   volume={52},
   ISSN={1545-4479},
   url={http://dx.doi.org/10.1146/annurev-fluid-010719-060214},
   DOI={10.1146/annurev-fluid-010719-060214},
   number={1},
   journal={Annual Review of Fluid Mechanics},
   publisher={Annual Reviews},
   author={Brunton, Steven L. and Noack, Bernd R. and Koumoutsakos, Petros},
   year={2020},
   month=jan, pages={477–508} }

@ARTICLE{Wang2023,
  author  = {Wang, X. and Ning, C. J. and Wang, W. Z. and others},
  title   = {Intelligent fusion method of multi-source aerodynamic data for flight tests},
  journal = {Acta Aerodynamica Sinica},
  volume  = {41},
  number  = {2},
  year    = {2023},
  pages   = {12--20},
  doi     = {10.7638/kqdlxxb-2021.0428}
}

@ARTICLE{Cui2024,
  author  = {Cui, R. F. and Wang, X. Y. and Liu, Z. and others},
  title   = {Aircraft multi-source data fusion method based on resampling weighting},
  journal = {Aeronautical Science and Technology},
  volume  = {35},
  number  = {7},
  year    = {2024},
  pages   = {111--119},
  doi     = {10.19452/j.issn1007-5453.2024.07.012}
}

@ARTICLE{Vaiuso2024,
  author  = {Vaiuso, A. and Immordino, G. and Righi, M. and Ronch, A.},
  title   = {Multi-fidelity {Bayesian} {Neural} {Network} for {Uncertainty} {Quantification} in {Transonic} {Aerodynamic} {Loads}},
  journal = {ArXiv},
  year    = {2024},
  eprint  = {abs/2407.05684}
}

@ARTICLE{Chen2025,
  author  = {Chen, L. W. and Qing, T. and Gao, W. H. and others},
  title   = {Application of {Multi-fidelity} {Data} {Fusion} {Methods} in the {Prediction} of {Aerodynamic} {Characteristics} for {Reusable} {Rockets}},
  journal = {Journal of Astronautics},
  volume  = {46},
  number  = {2},
  year    = {2025},
  pages   = {310--319},
  doi     = {10.3873/j.issn.1000-1328.2025.02.009}
}

@ARTICLE{Lin2024,
  author  = {Lin, F. and Hai, C. L. and Mei, L. Q.},
  title   = {Multi-source aerodynamic data fusion modeling with {XGBoost}},
  journal = {Acta Aerodynamica Sinica},
  volume  = {42},
  number  = {7},
  year    = {2024},
  pages   = {27--35},
  doi     = {10.7638/kqdlxxb-2023.0066}
}

@ARTICLE{He2020,
  author  = {He, L. and Qian, W. and Zhao, T. and Wang, Q.},
  title   = {Multi-fidelity aerodynamic data fusion with a deep neural network modeling method},
  journal = {Entropy},
  volume  = {22},
  number  = {9},
  year    = {2020},
  pages   = {1022},
  doi     = {10.3390/e22091022}
}

@ARTICLE{Nils2020,
  author  = {Thuerey, N. and Wei{\ss}enow, K. and Prantl, L. and Hu, X.},
  title   = {Deep {Learning} {Methods} for {Reynolds-Averaged} {Navier}--{Stokes} {Simulations} of {Airfoil} {Flows}},
  journal = {AIAA Journal},
  volume  = {58},
  number  = {1},
  year    = {2020},
  pages   = {25--36},
  doi     = {10.2514/1.J058291}
}

@ARTICLE{Ning2024,
  author  = {Ning, C. and Zhang, W.},
  title   = {{MHA-Net}: multi-source heterogeneous aerodynamic data fusion neural network embedding reduced dimension features},
  journal = {Aerosp Sci Technol},
  volume  = {145},
  year    = {2024},
  pages   = {108908}
}

@INPROCEEDINGS{Azad2021,
  author  = {Azad, R. and Fayjie, A. and Kauffmann, C. and others},
  title   = {On the texture bias for few-shot cnn segmentation},
  booktitle = {Proceedings of the IEEE/CVF Winter Conference on Applications of Computer Vision},
  year    = {2021},
  pages   = {2674--2683}
}

@ARTICLE{Vaswani2017,
  author  = {Vaswani, A. and Shazeer, N. and Parmar, N. and others},
  title   = {Attention is {All} you {Need}},
  journal = {Neural Information Processing Systems},
  year    = {2017}
}

@ARTICLE{Dong2025,
  author  = {Dong, W. and Wang, X. and Han, D. and Lin, Q.},
  title   = {Unsteady aerodynamic modeling and analysis of aircraft model in multi-dof coupling maneuvers at high angles of attack with attention mechanism},
  journal = {Chinese Journal of Aeronautics},
  volume  = {38},
  number  = {6},
  year    = {2025},
  pages   = {103444},
  doi     = {10.1016/j.cja.2025.103444}
}

@INPROCEEDINGS{WangY2025,
  author  = {Wang, Y. and Zhang, P. and Liu, Y. and others},
  title   = {Aerodynamic {Coefficients} {Prediction} via {Cross-Attention} {Fusion} and {Physical-Informed} {Training}},
  booktitle = {Proceedings of the AAAI Conference on Artificial Intelligence},
  volume  = {39},
  year    = {2025},
  pages   = {869--876},
  doi     = {10.1609/aaai.v39i1.32071}
}

@ARTICLE{Jones2001,
  author  = {Jones, D.},
  title   = {A taxonomy of global optimization methods based on response surfaces},
  journal = {Journal of Global Optimization},
  volume  = {21},
  year    = {2001},
  pages   = {345--383}
}

@ARTICLE{Han2016,
  author  = {Han, Z.},
  title   = {Kriging surrogate model and its application to design optimization: A review of recent progress},
  journal = {ACTA AERONAUTICA ET ASTRONAUTICA SINICA},
  volume  = {37},
  number  = {11},
  year    = {2016},
  pages   = {3197--3225}
}

@article{WANG2025111707,
title = {Multisource aerodynamic data reconstruction method using an enhanced multifidelity neural network},
journal = {Engineering Applications of Artificial Intelligence},
volume = {159},
pages = {111707},
year = {2025},
issn = {0952-1976},
doi = {https://doi.org/10.1016/j.engappai.2025.111707},
url = {https://www.sciencedirect.com/science/article/pii/S0952197625017099},
author = {Xu Wang and Huailu Li and Haitao Lin and Hui Tang and Weiwei Zhang}
}

@article{2022_Li,
author = {Kai Li and Jiaqing Kou and Weiwei Zhang},
title = {Deep Learning for Multifidelity Aerodynamic Distribution Modeling from Experimental and Simulation Data},
journal = {AIAA Journal},
year = {2022},
volume = {60},
publisher = {American Institute of Aeronautics and Astronautics (AIAA)},
month = {apr},
url = {https://doi.org/10.2514/1.j061330},
number = {7},
pages = {4413--4427},
doi = {10.2514/1.j061330}
}

@InProceedings{Fang2025,
author="Fang, Shiwei
and Xiang, Yu
and Zhang, Jun
and Wang, Wenyong",
editor="Hadfi, Rafik
and Anthony, Patricia
and Sharma, Alok
and Ito, Takayuki
and Bai, Quan",
title="A Novel Geometric-Encoded and Feature-Fused Model for Pressure Distribution Prediction on Airfoils",
booktitle="PRICAI 2024: Trends in Artificial Intelligence",
year="2025",
publisher="Springer Nature Singapore",
address="Singapore",
pages="134--146"
}

@article{ZHOU2025128242,
title = {Hybrid CNN-transformer network with frequency-aware fusion for efficient single image super-resolution},
journal = {Expert Systems with Applications},
volume = {288},
pages = {128242},
year = {2025},
issn = {0957-4174},
doi = {https://doi.org/10.1016/j.eswa.2025.128242},
url = {https://www.sciencedirect.com/science/article/pii/S0957417425018615},
author = {Ying Zhou and Zhichao Zheng and Quansen Sun and Rui Chen and Li Wang and Linghui Ge}
}

@article{XU2022,
title = "Comparison of machine learning data fusion methods applied to aerodynamic modeling of rocket first stage with grid fins",
author = "Chenzhou Xu and Tao Du and Zhonghua Han and Bowen Zan and Yu Mou and Jinze Zhang",
note = "Publisher Copyright: {\textcopyright} 2022 Zhongguo Kongqi Dongli Yanjiu yu Fazhan Zhongxin. All rights reserved.",
year = "2022",
month = jun,
doi = "10.11729/syltlx20210154",
volume = "36",
pages = "79--92",
journal = "Shiyan Liuti Lixue/Journal of Experiments in Fluid Mechanics",
issn = "1672-9897",
publisher = "Zhongguo Kongqi Dongli Yanjiu yu Fazhan Zhongxin",
number = "3"
}

@article{XU2024,
title = {Expert's experience-informed hierarchical kriging method for aerodynamic data modeling},
journal = {Engineering Applications of Artificial Intelligence},
volume = {133},
pages = {108490},
year = {2024},
issn = {0952-1976},
doi = {https://doi.org/10.1016/j.engappai.2024.108490},
url = {https://www.sciencedirect.com/science/article/pii/S0952197624006481},
author = {Chen-Zhou Xu and Zhong-Hua Han and Bo-Wen Zan and Ke-Shi Zhang and Gong Chen and Wen-Zheng Wang}
}

@techreport{cook1979,
  author      = {Cook, P. H. and McDonald, M. A. and Firmin, M. C. P.},
  title       = {Aerofoil {RAE} 2822: Pressure Distribution and Boundary Layer and Wake Measurements},
  institution = {Advisory Group for Aerospace Research and Development (AGARD)},
  year        = {1979},
  type        = {AGARD Advisory Report},
  number      = {AR-138},
  pages       = {A6-1--A6-77},
  note        = {Experimental Data Base for Computer Program Assessment, Appendix A6}
}

@article{HU2022108369,
title = {Aerodynamic data predictions based on multi-task learning},
journal = {Applied Soft Computing},
volume = {116},
pages = {108369},
year = {2022},
issn = {1568-4946},
doi = {https://doi.org/10.1016/j.asoc.2021.108369},
url = {https://www.sciencedirect.com/science/article/pii/S1568494621011418},
author = {Liwei Hu and Yu Xiang and Jun Zhang and Zifang Shi and Wenzheng Wang}
}

@article{HU2023108198,
title = {Flow field modeling of airfoil based on convolutional neural networks from transform domain perspective},
journal = {Aerospace Science and Technology},
volume = {136},
pages = {108198},
year = {2023},
issn = {1270-9638},
doi = {https://doi.org/10.1016/j.ast.2023.108198},
url = {https://www.sciencedirect.com/science/article/pii/S1270963823000950},
author = {Jiawei Hu and Weiwei Zhang}
}

@article{Liao2021,
  author  = {Liao, Peng and Song, Wei and Du, Peng and Zhao, Hang},
  title   = {Multi-fidelity convolutional neural network surrogate model for aerodynamic optimization based on transfer learning},
  journal = {Physics of Fluids},
  volume  = {33},
  number  = {12},
  pages   = {127121},
  year    = {2021},
  month   = {dec},
  issn    = {1070-6631},
  doi     = {10.1063/5.0076538},
  url     = {https://doi.org/10.1063/5.0076538}
}

@article{Bhatnagar2019,
  author = {Bhatnagar, Saakaar and Afshar, Yaghoub and Pan, Shaowu and Duraisamy, Karthik and Kaushik, Shailendra},
  title = {Prediction of aerodynamic flow fields using convolutional neural networks},
  journal = {Computational Mechanics},
  volume = {64},
  number = {2},
  pages = {525--545},
  year = {2019},
  doi = {10.1007/s00466-019-01740-0}
}

@article{WU2022470,
title = {A generative deep learning framework for airfoil flow field prediction with sparse data},
journal = {Chinese Journal of Aeronautics},
volume = {35},
number = {1},
pages = {470-484},
year = {2022},
issn = {1000-9361},
doi = {https://doi.org/10.1016/j.cja.2021.02.012},
url = {https://www.sciencedirect.com/science/article/pii/S1000936121000728},
author = {Haizhou WU and Xuejun LIU and Wei AN and Hongqiang LYU}
}

@article{WANG202262,
title = {An inverse design method for supercritical airfoil based on conditional generative models},
journal = {Chinese Journal of Aeronautics},
volume = {35},
number = {3},
pages = {62-74},
year = {2022},
issn = {1000-9361},
doi = {https://doi.org/10.1016/j.cja.2021.03.006},
url = {https://www.sciencedirect.com/science/article/pii/S1000936121000662},
author = {Jing WANG and Runze LI and Cheng HE and Haixin CHEN and Ran CHENG and Chen ZHAI and Miao ZHANG}
}

@article{Zuo2023,
  title={Fast simulation of airfoil flow field via deep neural network},
  author={Kuijun Zuo and Zhengyin Ye and Shuhui Bu and Xianxu Yuan and Weiwei Zhang},
  journal={Aerospace Science and Technology},
  year={2023},
  url={https://api.semanticscholar.org/CorpusID:266052708}
}

@article{Kennedy2000,
    author = {Kennedy, MC and O'Hagan, A},
    title = {Predicting the output from a complex computer code when fast approximations are available},
    journal = {Biometrika},
    volume = {87},
    number = {1},
    pages = {1-13},
    year = {2000},
    month = {03},
    issn = {0006-3444},
    doi = {10.1093/biomet/87.1.1},
    url = {https://doi.org/10.1093/biomet/87.1.1},
    eprint = {https://academic.oup.com/biomet/article-pdf/87/1/1/590577/870001.pdf}
}

@article{Perdikaris2017,
  title={Nonlinear information fusion algorithms for data-efficient multi-fidelity modelling},
  author={Perdikaris, Paris and Raissi, Maziar and Gartland, CP and Karniadakis, George Em},
  journal={Proceedings of the Royal Society A: Mathematical, Physical and Engineering Sciences},
  volume={473},
  number={2198},
  pages={20160751},
  year={2017},
  publisher={The Royal Society Publishing},
  doi={10.1098/rspa.2016.0751}
}

@article{Forrester2007,
    author = {Forrester, Alexander I.J and Sóbester, András and Keane, Andy J},
    title = {Multi-fidelity optimization via surrogate modelling},
    journal = {Proceedings of the Royal Society A: Mathematical, Physical and Engineering Sciences},
    volume = {463},
    number = {2088},
    pages = {3251-3269},
    year = {2007},
    month = {10},
    issn = {1364-5021},
    doi = {10.1098/rspa.2007.1900},
    url = {https://doi.org/10.1098/rspa.2007.1900},
    eprint = {https://royalsocietypublishing.org/rspa/article-pdf/463/2088/3251/713886/rspa.2007.1900.pdf},
}

@article{JeromeSacks1989,
 ISSN = {08834237, 21688745},
 URL = {http://www.jstor.org/stable/2245858},
 author = {Jerome Sacks and William J. Welch and Toby J. Mitchell and Henry P. Wynn},
 journal = {Statistical Science},
 number = {4},
 pages = {409--423},
 publisher = {Institute of Mathematical Statistics},
 title = {Design and Analysis of Computer Experiments},
 urldate = {2026-03-11},
 volume = {4},
 year = {1989}
}

@article{Han2012,
  title={Hierarchical Kriging model for variable-fidelity surrogate modeling},
  author={Han, Zhong-Hua and G{\"o}rtz, Stefan},
  journal={AIAA Journal},
  volume={50},
  number={9},
  pages={1885--1896},
  year={2012},
  publisher={American Institute of Aeronautics and Astronautics}
}

@inproceedings{Wi2025,
author = {Wi, Hyowon and Choi, Jeongwhan and Park, Noseong},
title = {Learning advanced self-attention for linear transformers in the singular value domain},
year = {2025},
isbn = {978-1-956792-06-5},
url = {https://doi.org/10.24963/ijcai.2025/730},
doi = {10.24963/ijcai.2025/730},
booktitle = {Proceedings of the Thirty-Fourth International Joint Conference on Artificial Intelligence},
articleno = {730},
numpages = {9},
location = {Montreal, Canada},
series = {IJCAI '25}
}

@inproceedings{Chen2026ArGEnT,
  title={ArGEnT: Arbitrary Geometry-encoded Transformer for Operator Learning},
  author={Wenqian Chen and Yuchen Fu and Michael Penwarden and Pratanu Roy and Panos Stinis},
  year={2026},
  url={https://api.semanticscholar.org/CorpusID:285541283}
}

@inbook{Cressie1993,
  title={Statistics for Spatial Data},
  author={Cressie, Noel A. C.},
  year={1993},
  publisher={John Wiley \& Sons},
  address={New York},
  pages={119--123}
}

@inbook{Rasmussen2006,
  title={Gaussian Processes for Machine Learning},
  author={Rasmussen, Carl Edward and Williams, Christopher K. I.},
  year={2006},
  publisher={The MIT Press},
  address={Cambridge, MA},
  pages={83--84}
}

\end{document}